%% file: main.tex
\documentclass[pdflatex,sn-basic]{sn-jnl}

\usepackage{graphicx}%
\usepackage{multirow}%
\usepackage[most]{tcolorbox}
\usepackage{amsmath,amssymb,amsfonts}%
\usepackage{amsthm}%
\usepackage{mathrsfs}%
\usepackage[title]{appendix}%
\usepackage{xcolor}%
\usepackage{textcomp}%
\usepackage{manyfoot}%
\usepackage{booktabs}%
\usepackage{algorithm}%
\usepackage{algorithmicx}%
\usepackage{algpseudocode}%
\usepackage{listings}%
\usepackage{subcaption}

\theoremstyle{thmstyleone}%
\theoremstyle{thmstyletwo}%

\theoremstyle{thmstylethree}%

\newcommand{\InputData}[1]{\item[\textbf{Input Data:}] #1}
\newcommand{\Hyperparameters}[1]{\item[\textbf{Hyperparameters:}] #1}
\newcommand{\OutputData}[1]{\item[\textbf{Output Data:}] #1}

\usepackage{amsthm}
\usepackage[framemethod=tikz]{mdframed}

\newmdtheoremenv[
  linecolor=gray,
  linewidth=1pt,
  topline=false,
  bottomline=false,
  leftline=true,
  rightline=false,
  backgroundcolor=gray!10,
  innerleftmargin=10pt,
  innertopmargin=5pt,
  innerbottommargin=5pt
]{definition}{Definition}[section]

\begin{document}

\title[Article Title]{The Urban Impact of AI: Modelling Feedback Loops in Next-Venue Recommendation}


\author*[1,2]{\fnm{Giovanni} \sur{Mauro}}\email{giovanni.mauro@sns.it}

\author[3]{\fnm{Marco} \sur{Minici}}\email{marco.minici@icar.cnr.it}

\author*[1,2]{\fnm{Luca} \sur{Pappalardo}}\email{luca.pappalardo@isti.cnr.it}

\affil[1]{ \orgname{ISTI-CNR}, \orgaddress{\street{via G. Moruzzi 1}, \city{Pisa}, \postcode{56124}, \state{Italy}}}

\affil[2]{ \orgname{Scuola Normale Superiore}, \orgaddress{\street{Piazza dei Cavalieri,7}, \city{Pisa}, \postcode{56126}, \state{Italy}}}

\affil[2]{ \orgname{ICAR-CNR}, \orgaddress{\street{Via Pietro Bucci 8/9c}, \city{Rende}, \postcode{87036}, \state{Italy}}}


\abstract{Next-venue recommender systems are increasingly embedded in location-based services, shaping individual mobility decisions in urban environments. 
While their predictive accuracy has been extensively studied, less attention has been paid to their systemic impact on urban dynamics. In this work, we introduce a simulation framework to model the human–AI feedback loop underpinning next-venue recommendation, capturing how algorithmic suggestions influence individual behaviour, which in turn reshapes the data used to retrain the models. 
Our simulations, grounded in real-world mobility data, systematically explore the effects of algorithmic adoption across a range of recommendation strategies.
We find that while recommender systems consistently increase individual-level diversity in visited venues, 
they may simultaneously amplify collective inequality by concentrating visits on a limited subset of popular places. This divergence extends to the structure of social co-location networks, revealing broader implications for urban accessibility and spatial segregation.
Our framework operationalizes the feedback loop in next-venue recommendation and offers a novel lens through which to assess the societal impact of AI-assisted mobility—providing a computational tool to anticipate future risks, evaluate regulatory interventions, and inform the design of ethic algorithmic systems.
}




\maketitle

\section{Introduction}
\label{sec:intro}

Recommender systems have become increasingly pervasive in urban life, seamlessly integrated into the online platforms that millions of individuals rely on to make daily decisions such as navigating cities and searching for accommodation \citep{pedreschi2024human, quijano2020recommender}. 
Among the most influential are next-venue recommenders (a.k.a. points-of-interest recommenders), which suggest places to visit -- such as restaurants, cafés, parks, shops, or cultural landmarks -- through widely used location-based services like Google Maps, Yelp, and Dianping. These platforms act as decision-making assistants, offering personalised suggestions in real time based on an individual's current location, mobility history, preferences, and contextual factors such as time of day or weather \citep{pappalardo2023future}.

The growing importance of next-venue recommenders stems from both their ubiquity and impact. 
As mobile devices and GPS-enabled apps become increasingly pervasive tools in our daily lives, we increasingly delegate micro-decisions (where to eat, shop, or spend leisure time) to algorithmic systems.
These recommendations are powered by machine learning techniques and deep learning architectures, which can model complex spatio-temporal patterns and predict human whereabouts with high accuracy~\citep{luca2021survey}.
By shaping when and where people move in the city, next-venue recommenders exert a subtle yet powerful influence on human movements, urban flows, and the popularity of venues. 
For example, a recent study shows that the recommender system implemented by Uber Eats led to significant improvements in consumer engagement and gross bookings, influencing the popularity of certain restaurants \citep{wang2025recommending}.\footnote{See also \url{https://www.gsb.stanford.edu/insights/better-way-make-recommendations-power-popular-platforms}} 

Although much research has focused on developing accurate next-location prediction algorithms~\citep{luca2021survey}, and recent work has begun to question their ability to generalise to unknown situations~\citep{luca2023trajectory}, their broader implications for urban dynamics remain largely unexplored~\citep{pedreschi2024human, pappalardo2023future, pappalardo2024survey}.
This gap in understanding is particularly pressing in light of recent regulatory developments. The European Union’s Digital Services Act introduces mandatory risk assessment requirements for very large online platforms and search engines, with a focus on assessing their impact on users and society.\footnote{\url{https://eur-lex.europa.eu/legal-content/EN/TXT/?uri=CELEX\%3A32022R2065}}
Among these platforms is Google Maps, one of the most influential systems that recommend physical venues, which is now officially subject to this regulatory scrutiny.


Understanding the urban impact of next-venue recommenders is challenging because their interaction with individuals creates an intricate feedback loop~\citep{pedreschi2024human}: each individual decision contributes to the formation of the big data that is used to train the recommender system, which in turn influences future decisions about where to go. 
This recursive process generates a continuously evolving cycle that can reshape the urban landscape.
Capturing this feedback loop in its entirety -- encompassing algorithmic recommendations, user responses, the parameters of machine learning models, and the retraining frequency -- is difficult, as it would require direct access to the online platforms delivering the recommender systems. At present, only the companies operating these platforms have the technical and legal means to observe and analyse the full feedback loop.
The Delegated Act detailing how vetted researchers can gain access to platform data under Article 40 of the Digital Services Act was only published in July 2025, and to date no scientific studies conducted by researchers outside the platforms have been able to experimentally analyse their impact.
Therefore, it is not surprising that most research so far has focused on mathematical formalisation of the feedback loop, rather than its empirical measurement \citep{pappalardo2024survey, chen2023bias}. 
While some progress has been made in this direction for other human-AI ecosystems -- such as social media, online retail, and generative AI~\citep{lesota2024oh, zhou2024sourceechochamberexploring, shumailov2024ai, mansoury2020feedback, jiang2019degenerate, sun2019debiasing, nguyen2014exploring, piao2023human, coppolillo2025algorithmic} -- no equivalent effort has yet been undertaken for feedback loops that involve next-venue recommender systems despite their growing influence on urban dynamics.
 
In this article, we propose a computational framework that models the feedback loop between individuals and next-venue recommender systems, with the goal of simulating its impact on individual and collective patterns of visits to public venues in a city.
Our simulation framework integrates multiple types of recommender systems -- from neighbourhood-based to deep learning methods -- trained on real-world data describing venue visits in New York City \textcolor{black}{and Tokyo}.
At the core of the framework is the venue selection mechanism, which governs how individuals decide where to go next: with a given probability, they follow a recommendation; otherwise, they make an autonomous choice, simulated using a mechanistic model of human mobility. 
We run the simulation over several weeks, allowing the feedback loop between individuals and the recommender system to unfold over time. As individuals interact with the system -- receiving recommendations, making choices, and visiting venues -- these interactions continually update the data used to retrain the recommender system, which in turn influences future recommendations. This dynamic process enables us to observe how different levels of algorithmic influence shape urban mobility patterns. Specifically, we study: \emph{(i)} the diversity of visits at the individual level, i.e. how varied each individual's venue choices are; \emph{(ii)} the variation in venue popularity at the collective level, i.e., how visits are distributed across all venues; and \emph{(iii)} the structure of the individuals' co-location network, i.e., how often individuals visit the same venues within a given time window.

Our results reveal that the impact of recommender systems varies across the various recommendation methods investigated, yet a consistent pattern emerges: at the individual level, we find an increased diversification in venue visits, while at the collective level, diversity decreases reflecting a rise in inequality in venue popularity. 
We further examine the tension between personalization and concentration, and find that the increase in individual diversity is largely driven by visits to already popular venues, ultimately making individuals' visitation patterns more similar to one another. 
This mirrors patterns observed in other human–AI ecosystems, where recommender systems improve individual choice but simultaneously amplify system-wide popularity bias.
Our study thus adds another piece to the broader mosaic of how recommender systems shape complex social systems, highlighting the structural consequences of algorithmic mediation across diverse domains.

The remainder of the paper is organised as follows.
Section~\ref{sec:related} reviews the literature on the impact of recommender systems on urban dynamics.
Section~\ref{sec:method} presents our modelling of the feedback loop and the simulation framework.
Section~\ref{sec:settings} outlines the experimental setup and Section~\ref{sec:results} reports the results of our simulations.
Section~\ref{sec:discussion} interprets our findings in the broader context of human–AI coevolution and human mobility.
Finally, Section~\ref{sec:conclusions} concludes the paper by discussing the limitations of our study and proposing directions for future research.

\section{Related Work}
\label{sec:related}

This section reviews key experimental studies that examine how recommendation systems impact urban dynamics.

A growing body of research investigates the impact of GPS-based navigation systems on the urban environment. 
\citet{arora2021quantifying} simulate the effects of Google Maps navigation in Salt Lake City, reporting average reductions of 6.5\% in travel time and 1.7\% in CO$_2$ emissions. These benefits are even more pronounced for individuals whose routes were altered by the route recommendations. \citet{cornacchia2022routing} use TomTom APIs to simulate navigation services in Milan, showing that adoption rates critically shape system-wide outcomes: emissions increase at very low or very high adoption levels but decrease when adoption stabilises around 50\%. 
\citet{perezprada2017managing} show that the widespread adoption of eco-routing during congestion can reduce CO$_2$ and NO$_x$ emissions by 10\% and 13\%, respectively. However, these gains come with trade-offs: NO$_x$ exposure increases by 20\%, travel times rise by 28\%, and vehicle concentration in downtown areas grows by 16\%. 
Two studies simulate routing systems in Florence, Rome, and Milan to show that prioritising diversity in routing can lead to more balanced traffic distribution, improved road network coverage, and reductions in CO$_2$ emissions~\citep{cornacchia2024alternative, cornacchia2023oneshot}.

Several studies examine how recommendation systems on house-renting platforms influence socio-economic dynamics, often reinforcing systemic inequalities. \citet{edelman2014digital} uncover racial disparities in New York City’s Airbnb market, reporting a 12\% revenue gap between Black and non-Black hosts, even after controlling for property characteristics and guest ratings. \citet{zhang2021frontiers} expand on this finding, showing that the AirBnB smart pricing algorithm can reduce these price disparities and improve revenues for black hosts, although it does not completely eliminate systemic bias.

Additional research investigates how ride-hailing and ride-sharing platforms influence urban mobility through algorithmic matching and incentive design.  \citet{bokanyi2020understanding} find that a ride recommender system prioritising lower-earning drivers in New York City would reduce disparities and increase driver earnings compared to prioritizing the closest available vehicle. \citet{afeche2023ridehailing} analyse passenger-driver matching and centralised dispatch algorithms, showing that while centralized control improves efficiency, it can restrict service access in less densely populated areas, exacerbating spatial inequities.

\paragraph{Position of our work}
A review of the literature reveals two major gaps \citep{pappalardo2024survey}. 
First, there are no existing studies that have examined the actual or potential impact of next-venue recommendation systems on urban dynamics. 
Second, although a growing body of research outside the urban domain has begun to formalise and model human–AI feedback loops~\citep{sun2019debiasing, mansoury2020feedback, jiang2019degenerate, nguyen2014exploring, shumailov2024ai, lesota2024oh, zhou2024sourceechochamberexploring, coppolillo2025algorithmic}, little attention has been paid to how these feedback loops manifest in urban environments or how they can be formally modelled.

A notable exception is the work by \citet{ensign2018runaway}, which shows how a feedback loop can distort the allocation of police resources, reinforce existing biases, and generate unrealistic crime distributions in a city. 
By introducing corrective mechanisms to prevent repeated deployments in the same areas, the authors demonstrate that even minor changes in algorithmic design can significantly influence urban outcomes. 
While this study represents a pioneering effort, the modelling of human–AI feedback loops and their effects on urban mobility and spatial behaviour remain largely unexplored.

Our work takes a first step toward addressing these gaps by explicitly modelling the human–AI feedback loop in the context of next-venue recommendation systems. We introduce an open-source simulation framework that enables the systematic evaluation of how different recommendation algorithms influence key aspects of urban dynamics, such as venue popularity and individual visitation patterns.

\section{Methodology}
\label{sec:method}

We design a simulation framework to model the human-AI feedback loop in the context of next-venue recommendation. 
In Section~\ref{sec:preliminaries}, we introduce the notation and definitions necessary to understand the simulation framework. Section~\ref{sec:simulation_framework} presents our modelling of the feedback loop and the simulation methodology. Section~\ref{sec:metrics} describes the metrics used to evaluate the impact of the feedback loop on individual and collective venue visitation patterns.

\input{method}

\section{Experimental settings}
\label{sec:settings}
In this section, we describe the experimental settings of our study.
Section~\ref{sec:benchmark} presents the recommendation systems evaluated in our simulations.
Section~\ref{sec:dataset} details the dataset on which our experiments are based, and Section~\ref{sec:parameters} outlines the values and configurations of the parameters used in the simulation framework.

\input{settings}

\section{Results}
\label{sec:results}
\input{experiments}

\section{Discussion}
\label{sec:discussion}

Our findings contribute to the growing body of research on human–AI interaction by extending the analysis of feedback loops to the domain of human mobility. 

Our main result is that the feedback loop leads to a divergence between individual and collective diversity for most recommendation systems. 
On the one hand, individuals distribute their visits more evenly across venues, resulting in increased individual diversity. 
On the other hand, the distribution of collective venue popularity becomes more concentrated, with a small set of venues attracting a disproportionate share of total visits -- a decreased collected diversity.
These two apparently contrasting mechanisms coexist because individual diversity rises mostly through visits to the same popular venues, which makes individual visitation patterns increasingly similar.

This individual-collective trade-off mirrors the patterns observed in other human-AI ecosystems \citep{pappalardo2024survey}. 
In online retail, for example, the use of recommender systems often increases individual exposure to niche items while amplifying popularity bias at the aggregate level, leading to market concentration around top-ranked products~\citep{lee2019recommender, fleder2009blockbuster}. 
A similar dynamic has been observed in the context of generative AI.
Online experiments show that while generative models can boost individual creativity, they often reduce diversity at the collective level~\citep{doshi2024generative}.
Additionally, when data produced by an AI model is fed back into the model for fine-tuning, the resulting feedback loop can lead to mode collapse, producing increasingly homogenized outputs at scale~\citep{shumailov2024ai}.
Our results suggest that the human-AI feedback loop produces this trade-off not only in digital content consumption but also in physical movement patterns within cities. 
This raises important concerns about the long-term effects of algorithmic mediation on urban accessibility, spatial inequality, and the viability of less popular or peripheral venues. 
In this regard, our study contributes to the emerging field of human-AI coevolution~\citep{pedreschi2024human}, which focuses on measuring and modelling feedback loops between humans and algorithms, and understanding how these interactions shape the dynamics of complex systems.

\textcolor{black}{
Another key finding of this study is that the extent of the individual-collective trade-off depend on the algorithmic architecture used to train the recommender systems within the feedback loop. 
Empirical research shows that different types of recommender systems introduce distinct forms of bias, with no universally accepted solution: for example, neighbourhood-based models tend to exhibit less popularity and exposure bias than deep learning approaches, while graph neural networks are prone to degree bias, disproportionately favouring highly connected nodes~\citep{chen2023bias, klimashevskaia2024survey, mansoury2019bias, lesota2021analyzing, elahi2021investigating, subramonian2024theoretical}.
Our findings are consistent with these observations: feedback loops with neighbourhood-based recommenders such as UserKNN and ItemKNN produce less pronounced biases than those with deep learning models such as MultiVAE. 
While our study does not offer a definitive explanation for the origins of these disparities, it adds evidence to the growing literature on how algorithmic design shapes fairness, diversity, and systemic outcomes in human-AI feedback loops.}

Our work also contributes to the field of human mobility~\citep{pappalardo2023future, barbosa2018human}, which seeks to analyse, predict, and model the fundamental patterns underlying human movement. 
In particular, our study challenges the core assumptions of mechanistic mobility models such as the Exploration and Preferential Return (EPR) model~\citep{song2010modelling} and its extensions \citep{barbosa2018human}, which typically assume that individuals make movement decisions autonomously, driven by internal rules balancing exploration and return.
However, in today’s algorithmically mediated urban environments, these decisions are often shaped by external influences (most notably, recommender systems) which embed individuals within a human-AI feedback loop that alters the dynamics of mobility itself. 
As our results show, such feedback loop has the potential to reshape venue visitation patterns, with cascading effects on venue popularity and social co-location.
These findings suggest that classical mobility models must be revised to explicitly incorporate algorithmic influence as a primary behavioural driver, enabling more realistic representations of human mobility in algorithmically guided cities. 


\section{Conclusion}
\label{sec:conclusions}
\input{conclusions}

\section*{Reproducibility and Code}

Our framework and analysis are fully reproducible, with all code available at \url{https://github.com/mauruscz/UrbanFeedbackLoop}. 

The simulations were executed on high-performance computing platforms with 64 CPU cores (AMD EPYC 7313 16-Core Processor) and $\sim$1.18 TB RAM.
While the simulations are computationally intensive, we implemented a highly parallelized and modular architecture to optimize performance and resource utilization. Experiments with neural recommenders, i.e., MultiVAE, LightGCN, BPRMF, have been conducted on a DGX Server equipped with 4 NVIDIA Tesla V100 GPU (32GB) and CUDA Version 12.2.

\backmatter


\bmhead{Authors' contributions} 
GM and MM implemented the code for simulations and the data analysis. GM, MM, and LP conceptualised the work and designed the figures. GM made the plots. GM, MM and LP wrote the paper. LP directed and supervised the research. All authors contributed to the scientific discussion, read and approved the paper.

\bmhead{Acknowledgements}
Giovanni Mauro has been supported by SoBigData.it—Strengthening the Italian RI for Social Mining and Big Data Analytics”, prot. IR0000013, avviso n. 3264 on 28/12/2021.

Marco Minici acknowledges partial support by $(i)$ the SERICS project (PE00000014) under the NRRP MUR program funded by the EU - NGEU and $(ii)$ ``SoBig-Data.it - Strengthening the Italian RI for Social Mining and Big Data Analytics" - Prot. IR0000013, avviso n. 3264 on 28/12/2021.

Luca Pappalardo has been supported by PNRR (Piano Nazionale di Ripresa e Resilienza) in the context of the research program 20224CZ5X4\_PE6\_PRIN 2022 “URBAI – Urban Artificial Intelligence” (CUP B53D23012770006), funded by European Union – Next Generation EU.

We thank Daniele Fadda for the invaluable help with the data visualisations.
We thank Andrea Frasson for his master's thesis work, which inspired this study. 
We thank Dino Pedreschi, Giuliano Cornacchia, Gabriele Barlacchi, Daniele Gambetta and Margherita Lalli for the useful discussions.

\bibliography{references}

\begin{appendices}
\input{appendix}
\end{appendices}

\end{document}

%% file: method.tex
\subsection{Preliminaries}
\label{sec:preliminaries}

Let $U$ be a set of individuals and $V$ a set of public venues. 
We define a dataset $D \subseteq U \times V \times T$ as an interaction matrix, where each tuple $(u, v, t) \in D$ represents a visit by individual $u \in U$ to venue $v \in V$ at time $t \in T$.  
The dataset describes all visits up to a time $T_{\text{max}}$, and tuples in it are sorted chronologically.
For each individual $u \in U$, we define $D_u = \{v_1, \dots, v_n\}$ as the time-ordered sequence of venues visited by $u$ in $D$, sorted chronologically. 
The dataset $D$ is partitioned into two disjoint subsets: a training set $D_{\text{train}}$, which contains all tuples $(u, v, t) \in D$ such that $t \leq T_{\text{train}}$, and a post-training set $D_{\text{post}}$, which contains all tuples such that $T_{\text{train}} < t \leq T_{\max}$.

Each venue $v \in V$ is associated with a categorical label provided by a function $\ell: V \rightarrow C$, where $C = \{c_1, \dots, c_z\}$ is a predefined set of venue categories (e.g., \emph{restaurant}, \emph{museum}, \emph{school}).
Venues are geolocated via a function $\pi: V \rightarrow [-90, 90] \times [-180, 180]$, which maps a venue $v \in V$ to its latitude and longitude coordinates. The geographical distance between two venues $v_i, v_j \in V$ is computed as the great-circle distance between their coordinates on the Earth’s surface.

We define a scoring function $\mathcal{R}_{\theta} : U \times V \times T \rightarrow [0,1]$ with parameters $\theta$ and trained on $D_{\text{train}}$. 
For any individual $u \in U$, venue $v \in V$, and time $t \in T$, the score $\mathcal{R}_{\theta}(u, v, t)$ represents the estimated probability that $u$ will visit $v$ as next location at a future time $t' > t$.
The recommender system, $\mathcal{A}$, is an operator that, for a given user $u$ and time $t$, ranks all candidate venues according to their scores $\mathcal{R}_{\theta}(u, v, t)$ and returns the top-$k$ venues from this ranking.


\begin{definition}[Next-venue recommendation problem]
Given a set of individuals $U$, a set of venues $V$, and a time $t$, the next-venue recommendation problem consists in predicting the venue $v \in V$ that an individual $u \in U$ is most likely to visit next. 
The goal of a next-venue recommender system $\mathcal{A}$ is to learn a function $R_\theta(u, v, t)$ that estimates the likelihood of future interactions and to return the top-$k$ venues with highest scores.
\label{def:problem}
\end{definition}

\subsection{Modeling and Simulating the Feedback Loop}
\label{sec:simulation_framework}

\begin{figure}
    \centering
    \includegraphics[width=\linewidth]{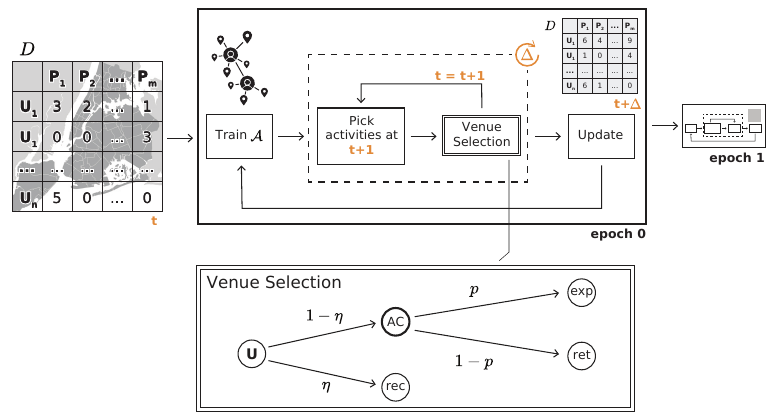}
    \caption{\textbf{Overview of the simulation framework modelling the feedback loop.}
    At each step of the simulation, an individual chooses between relying on recommendations (with probability $\eta$) or an autonomous choice (AC) (with probability $1-\eta$).  
    AC is split between exploration (with probability $p$) and preferential return (with probability $1 - p$), where $p$ is a fixed parameter equals to $\rho \cdot S^{-\gamma}$ where $S$ is the number of distinct previously visited locations.
    Initially, a recommender system $\mathcal{A}$ is trained on the history of venue visits. 
    At each timestep $t+1$, individuals' next venues are selected, and the dataset is updated accordingly. After $\Delta$ timesteps, the recommender is retrained, allowing the feedback loop between individual decisions and algorithmic suggestions to evolve over time.}
    \label{fig:schema}
\end{figure}

Given the sets $U$, $V$, the datasets $D$, $D_{\text{train}}$ and $D_{\text{post}}$, the scoring function $\mathcal{R}_\theta$ and the recommender system $\mathcal{A}$, we model the feedback loop to simulate the generation of visits to venues in $V$ by individuals in $U$. 

\textcolor{black}{We assume a scenario where individuals have a general intent or agenda (e.g., visiting a restaurant or bar) and then choose a specific venue based on either personal knowledge or the algorithmic suggestions provided by recommender system $\mathcal{A}$.} 
As the simulation of the feedback loop goes by, a dataset $\mathcal{S}$ is generated, which contains the simulated visits of the individuals to the venues.
Figure~\ref{fig:schema} provides an overview of the framework that models the feedback loop, illustrating the key components and the flow of interactions between individuals and the recommender system.

The simulation for each individual $u \in U$ is outlined in Algorithm~\ref{alg:simulation_framework}. 
The process begins with training the recommender system $\mathcal{A}$ on the dataset $D_{\text{train}}$ (line 1). 
The simulation then iterates over each visit tuple $(u, v, t) \in D_{\text{post}}$ (line 5), simulating user mobility in three sequential steps: category selection (line 6), where the category of the next venue is taken from $D_{\text{post}}$; venue selection (line 7), where the specific venue is selected based on the chosen category and the individual's decision to use or not the recommender system; and algorithm retraining (line 10), which updates $\mathcal{A}$ every $\Delta$ time units to incorporate newly simulated data in $\mathcal{S}$.

\begin{algorithm}[htb!]
\caption{Simulation Framework that models the Feedback Loop}
\label{alg:simulation_framework}
\begin{algorithmic}[1]
\InputData{Dataset of visits $D$, Training dataset $D_{\text{train}}$, Observation dataset $D_{\text{post}}$. All datasets are sorted by time.}
\Hyperparameters{Adoption rate $\eta$, Retraining frequency $\Delta$}
\OutputData{Dataset $\mathcal{S}$ of simulated visits}

\State $\mathcal{A} \leftarrow \text{AlgorithmTraining}(D_{\text{train}})$ \Comment{Train the recommender system}
\State $(u, v, t) \leftarrow \text{last}(D_{\text{train}})$ \Comment{Last tuple of the training dataset}
\State $t_{\text{last\_training}} \leftarrow t$ \Comment{Timestamp of the last training}
\State $\mathcal{S} \leftarrow \varnothing$ 

\For{$(u, v_{\text{current}}, t)$ in $D_{\text{post}}$} \Comment{Iterate over the visits}
    \State $c \leftarrow \text{CategorySelection}(v_{\text{current}})$ \Comment{Select the category of the next venue}
    \State $v_{\text{next}} \leftarrow \text{VenueSelection}(u, v_{\text{current}}, t, \mathcal{A}, \eta)$ \Comment{Select the next venue}
    \State $\mathcal{S} \leftarrow \mathcal{S} \cup \{(u, v_{\text{next}}, t)\}$ \Comment{Add the simulated visit to the dataset}
    \If{$t - t_{\text{last\_training}} > \Delta$} \Comment{Check if it's time to retrain the model}
        \State $\mathcal{A} \leftarrow \text{AlgorithmTraining}(D_{\text{train}} \cup \mathcal{S})$ \Comment{Retrain the recommender system}
        \State $t_{\text{last\_training}} \leftarrow t$ \Comment{Update last training time}
    \EndIf
\EndFor
\end{algorithmic}
\end{algorithm}

\subsubsection*{Category selection} 
For each visit by an individual $u$ in $D_{\text post}$, we observe the actual venue $v \in V$ visited by $u$ and determine the corresponding category $c = \ell(v)$.
{\color{black}Using the actual category of the visited venue, as observed in the data, serves as a proxy for the underlying intention or agenda of the individual $u$.}

\subsubsection*{Venue Selection}
Given the target category $c$ representing the type of venue to be visited, the individual $u \in U$ must select a specific venue of that category. The venue selection process is detailed in Algorithm~\ref{alg:venue_selection}.

The first step involves deciding whether to rely on the recommender system $\mathcal{A}$ or on personal knowledge (line 4 of Algorithm \ref{alg:venue_selection}). This decision reflects the user's choice between following algorithmic suggestions or acting autonomously. It is modelled as a Bernoulli trial with success probability $\eta \in [0,1]$, referred to as the adoption rate, which quantifies the likelihood that $u$ accepts the recommendation.
The parameter $\eta$ plays a central role in the simulation, controlling the degree to which the user behaviour is shaped by the recommender. If the Bernoulli trial succeeds, $u$ follows the recommendation (recommendation-based choice, line 5); otherwise, $u$ selects a venue independently (autonomous choice, lines 7--10).

\begin{algorithm}[htb!]
\caption{Venue Selection for Individual $u$}
\label{alg:venue_selection}
\begin{algorithmic}[1]
\InputData{Venue set $V$, Current venue $v_{\text{current}} \in V$, Target category $c$, User history $D_u$, Training dataset $D_{\text{train}}$, Simulated dataset $\mathcal{S}$, Jump length distribution $\mathcal{P}_{\text{jump}}$}
\Hyperparameters{Adoption rate $\eta$, Exploration probability $p$}
\OutputData{Selected venue $v_{\text{next}}$}

\State $r \leftarrow \text{Sample}(\mathcal{P}_{\text{jump}})$ \Comment{Sample a jump length from the empirical distribution}
\State $r^* \leftarrow \text{Median}(\mathcal{P}_{\text{jump}})$ \Comment{Median of jump distribution}
\State $V_{c, r}^{(v_{\text{\tiny current}})} \leftarrow \{v \in V \mid \ell(v) = c \land \text{dist}(v_{\text{current}}, v) \le r \}$ \Comment{Candidate venues}

\If{$\text{Bernoulli}(\eta)$} 
    \State $v_{\text{next}} \leftarrow \mathcal{A}(u, V_{c, r}^{(v_{\text{\tiny current}})})$ \Comment{Select venue using recommender system}
\Else 
    \If{$\text{Bernoulli}(1 - p)$} \Comment{Return to a previously visited venue}
        \State $v_{\text{next}} \leftarrow \text{PreferentialReturn}(D_u, c)$ 
    \Else \Comment{Explore a new venue}
        \State $v_{\text{next}} \leftarrow \text{Explore}(V_{c, r}^{(v_{\text{\tiny current}})}, D_u, r^*)$ 
    \EndIf
\EndIf
\end{algorithmic}
\end{algorithm}

\paragraph{Recommendation-based choice}
If an individual $u$ follows the suggestion of the recommender system $\mathcal{A}$ (line 5 of Algorithm \ref{alg:venue_selection}), a set of candidate venues $V_{c, r}^{(v_{\text{\tiny current}})} \subseteq V$ is retrieved (line 3, Algorithm \ref{alg:venue_selection}). 
This set includes all venues belonging to the target category $c$ and located within a radius $r$ of the current location of the individual $v_{\text{current}}$.

The radius $r$ is sampled from the empirical distribution of jump lengths observed in $D$, reflecting the tendency of individuals to prefer short-range displacements~\citep{gonzalez2008understanding}. 
This distribution is obtained by enumerating all pairs of consecutive venues $(v_{i-1}, v_i)$ in the visit sequences $D_u$ for each user $u \in U$, and computing the great-circle distance between $v_{i-1}$ and $v_i$ using their geographic coordinates.

Each venue $v \in V_{c, r}^{(v_{\text{\tiny current}})}$ is assigned a score $\mathcal{R}_\theta(u, v, t)$. 
The resulting score set $S$ is normalized using min-max scaling, yielding the set of normalized scores $\hat{S}$. 
The recommender system $\mathcal{A}$ then returns the top-$k$ venues in $L \in V_{c, r}$ based on $\hat{S}$. Finally, $u$ selects a venue $v_{\text{next}} \in L$ with probability proportional to its normalised score $\hat{S}(v)$.

\paragraph{Autonomous choice} 
If the individual $u$ moves independently of the recommender system $\mathcal{A}$ (line 7 of Algorithm \ref{alg:venue_selection}), we model their choice of venue using a modified version of the Exploration and Preferential Return (EPR) model, a well-established framework for human mobility introduced by~\cite{song2010modelling}.
\textcolor{black}{The EPR model captures the fundamental mechanism of human mobility as a trade-off between returning to previously visited locations and exploring new ones. 
It is a foundational model in the study of human mobility~\citep{barbosa2018human}and its flexibility makes it a suitable basis for modelling autonomous mobility decisions in our framework.}

{\color{black}Based on the EPR model}, at each decision point, we let the individual $u$ choose between: 
\begin{itemize} 
    \item \textbf{Preferential return} (line 8, Algorithm \ref{alg:venue_selection}): with probability $1 - p$, the individual selects a venue from their set of previously visited locations, with probability proportional to the number of times each venue has been visited by $u$.
    
    \item \textbf{Exploration} (line 10, Algorithm \ref{alg:venue_selection}): with probability $p$, the individual selects a new (unvisited) venue located within a radius $r$, where $r$ is sampled from the empirical distribution of jump lengths observed in $D$. 
    This radius-based sampling reflects the heavy-tailed nature of human travel patterns, where individuals occasionally make long jumps but typically explore within a constrained geographic range~\citep{gonzalez2008understanding}. Among the venues within this radius, the probability of selecting a specific venue $v$ is proportional to its relevance.
    \textcolor{black}{The radius-based sampling in the exploration phase reflects the heavy-tailed nature of human mobility~\citep{gonzalez2008understanding} and aligns with extensions of the EPR model \citep{barbosa2018human, pappalardo2015returners, pappalardo2016human}. 
    We assume that individuals favour dense or attractive areas, rather than choosing uniformly at random within the radius, as suggested by~\cite{pappalardo2015returners, pappalardo2016human}.}
    The relevance of a venue $v \in V$ is defined as the number of other venues located within a fixed radius $r^*$ from $v$, where $r^*$ is set to the median of the jump length distribution computed on $D$.
    Intuitively, a venue is more relevant if it lies in a dense area where many other venues fall within this typical range, making it more likely to be part of users' mobility patterns.
    Formally, the relevance is given by $rel(v) = | \{ v' \in V \setminus \{v\} | d(v, v') \leq r^* \}|$.
\end{itemize}

{\color{black}If a user lacks enough venues of the target category $c$ within the sampled radius $r$, a fallback mechanism ensures a valid venue is selected.
First, the mechanism searches for a venue in a broader category within the same radius.
For example, if the target category is \emph{Sushi Restaurant}, the system will attempt to find any venue under the broader \emph{Food} category.
If no venue is found within this broader category, the individual is assigned the geographically nearest venue of the original target category, regardless of distance.
Similarly, in the return phase, if the individual has not previously visited any venue in the target category, we look for one in the broader category.
If that also fails, the individuals explores, visiting a new venue.}


\subsubsection*{Algorithm retraining}
The recommender system $\mathcal{A}$ is periodically retrained during the simulation at fixed intervals of $\Delta$ time units (line 10, Algorithm \ref{alg:simulation_framework}).
At each retraining step, the model is updated using an enriched dataset $D_{\text{train}} \cup \mathcal{S}$, consisting of the original training set $D_{\text{train}}$ combined with the set $\mathcal{S}$ of simulated visits generated up to that point.
In practice, this step corresponds to updating the parameters $\theta$ of the scoring function $\mathcal{R}_\theta$ based on this enriched dataset.

This retraining procedure allows $\mathcal{A}$ to iteratively refine its understanding of individual preferences by incorporating both historical and simulated behavioural data. In doing so, it captures the dynamic nature of the human–AI feedback loop, where algorithmic outputs influence human behaviour, which in turn shapes future recommendations.

\subsection{Metrics}
\label{sec:metrics}
We assess the impact of the human–AI feedback loop on urban dynamics from two complementary perspectives: one focused on venues and the other on individuals.

\subsubsection*{Venue perspective}
We evaluate the inequality in the distribution of visits across venues using the Gini coefficient. 
This measure captures the extent to which visits are concentrated on a subset of venues: a Gini coefficient of $0$ indicates perfect equality (all venues receive the same number of visits), while higher values indicate increasing concentration of visits on fewer venues.
Our use of the Gini coefficient is motivated by studies showing that recommender systems can amplify the inequality in the distribution of consumed items~\citep{pedreschi2024human, pappalardo2024survey,
boratto2021connecting, elahi2021investigating, lee2019recommender}. 
In our context, the Gini coefficient quantifies to what extent algorithmic recommendations direct individuals towards a limited number of venues, increasing their popularity while marginalising others.

Given a dataset of visits $D$, we compute the collective Gini coefficient as:

\begin{equation} G = \frac{1}{|V|} \left( |V| + 1 - 2 \cdot \frac{\sum_{i=1}^{|V|} (|V| + 1 - i)  x_i}{\sum_{v \in V} x_i} \right), \label{eq:gini} \end{equation}

\noindent where $x_i = |{(u, v, t) \in D : v = v_i}|$ is the total number of visits to the venue $v_i \in V$, and the values $x_i$ are sorted in ascending order.

To complement this collective view, we also measure inequality from the perspective of individual behaviour. For each individual $u \in U$, we calculate an individual Gini coefficient $G_u$, applying Equation \ref{eq:gini} to $D_u$ (the visits made by $u$). 
That is, we compute the inequality in how frequently each venue is visited by the individual.
The average individual Gini coefficient is then given by:

\begin{equation} \overline{G} = \frac{1}{|U|} \sum_{u \in U} G_u, \end{equation}

\noindent which captures the typical inequality in venue usage across individuals.

\subsubsection*{Individual perspective}
From the individual perspective, we construct a co-location network in which nodes represent individuals and edges indicate co-presence at the same venue within a specified time window. Intuitively, this network captures how individuals encounter one another through shared spatial activity in the city.

Formally, let $D = {(u_i, v_j, t_k)}$ be the dataset of individual–venue visits. Given a time window $[t_1, t_2]$, the co-location network is defined as an undirected graph $\mathcal{G} = (U', E)$, where $U' \subseteq U$ is the set of individuals active in that time window (i.e., who move at least once), and $(u_m, u_n) \in E$ if and only if individuals $u_m$ and $u_n$ visited the same venue $v_j$ within the time window. That is $
\exists\, (u_m, v_j, t_k),\ (u_n, v_j, t_l) \in D$ such that $t_k, t_l \in [t_1, t_2]$.


To investigate the presence of hierarchical structure in the social interactions encoded by the co-location network, we analyse the tendency of high-degree nodes to form a densely interconnected core (rich club). 
There is substantial evidence showing that rich-club structures can influence the dynamics of complex systems \citep{pedreschi2022temporal, bertagnolli2022functional, berahmand2018effect, colizza2006detecting}, such as fostering socio-economic development \citep{kang2022measuring} and shaping the spatial organisation of cities \citep{jia2022dynamical}.

Following the method proposed by \citet{zhou2004rich}, we identify the set \( U_{\text{rich}} \) as the \( h \) nodes with the highest degree in \( G \), and compute the rich-club density as the ratio of the number of actual links among \( U_{\text{rich}} \) to the maximum possible number of links in a complete subgraph of \( h \) nodes, i.e. \( \frac{h(h-1)}{2} \).



We also examine how the degree distribution of social interactions may be affected by the influence of the recommender system. 
Given that the degree distribution $P(k)$ in real-world networks often exhibits heavy-tailed behaviour, we fit a linear regression to the log-log plot of $P(k)$ derived from the co-location network $\mathcal{G}$, and report the absolute value of the slope of the fitted line, denoted by $\alpha$. 
Larger values of $\alpha$ correspond to a steeper decline, reflecting a distribution closer to exponential decay, whereas smaller values reflect a more uniform structure. 
We estimate $P(k)$ using the empirical degree distribution, defined as
$P(k) = \frac{\lvert \{ u \in U^{\prime} : \deg(u) = k \} \rvert}{\lvert U^{\prime} \rvert}$,
where $\deg(\cdot)$ denotes the degree of a node in $\mathcal{G}$.



%% file: settings.tex
\subsection{Benchmark recommenders}
\label{sec:benchmark}
Various recommender systems have been designed to assist users in discovering the most relevant and valuable content \citep{shapira2022recommender, ricci2021recommender}. 
Among the various approaches, collaborative filtering represents one of the most widely adopted families of algorithms, as it relies solely on binary feedback indicating whether a user has interacted with a particular item \citep{koren2021advances}. 
These techniques have also been successfully applied to the recommendation of venues in urban settings \citep{rahmani2022role}, consistently achieving high accuracy across benchmark datasets for next-venue recommendation.

We benchmark several recommendation models spanning classical neighbours-based methods, matrix factorization techniques, deep generative models, geographical recommenders, and approaches based on graph neural networks. 
The list of models included in our study is detailed below:

\begin{itemize}
\item \textbf{User-KNN} \citep{aiolli2013efficient}: Neighbourhood-based method that recommends items based on the preferences of users with similar interaction histories;

\item \textbf{Item-KNN} \citep{aiolli2013efficient}: Similar to User-KNN, but 
recommends items that are similar to those the user has previously interacted with;

\item \textbf{Matrix Factorization (MF)} \citep{sarwar2000application}: Model-based approach that decomposes the user--item interaction matrix into latent user and item feature vectors, enabling prediction of missing interactions;

\item \textbf{Bayesian Personalized Ranking (BPRMF)} \citep{rendle2009bpr}: Extension of matrix factorization that optimizes for pairwise ranking by assuming users prefer observed items over unobserved ones, and learning to rank accordingly;

\item \textbf{Multinomial Variational Autoencoder (MultiVAE)} \citep{liang2018variational}: Generative model based on deep learning that captures user preferences by learning a probabilistic latent space using variational inference;
\item \textbf{Light Graph Convolutional Network (LightGCN)} \citep{he2020lightgcn}: Recommendation system based on graph neural networks that propagates user and item embeddings over a user--item bipartite graph;
\item \textbf{PGN} \citep{sanchez2022travelers}: Hybrid approach that averages the scores of User-KNN, a popularity-based recommender, and a geographical model that suggests venues near the centroid of the user's previously visited locations. 
\end{itemize}

As different models exhibit varying levels of bias in next-venue recommendation \citep{rahmani2022unfairness}, 
and algorithmic bias is directly linked to the long-term effects of feedback loops \citep{mansoury2020feedback, chaney2018algorithmic},
our comprehensive evaluation is designed to capture the different behavioural shifts and inequality patterns that may emerge from each recommendation strategy over time.
Details about the hyperparameters and the training procedure of the recommender systems are in Appendix \ref{appendix:recsys}.
As a sanity check, we report the performance of all the recommenders trained on $D_{\text{train}}$ and tested on $D_{\text{post}}$ in terms of $HitRate@20$ and $mRR@20$, see Table \ref{tab:performance} in Appendix \ref{appendix:recsys}.

\subsection{Dataset}
\label{sec:dataset}
We use a Foursquare dataset containing mobility data from New York City, collected over a period of 304 days -- from April 3, 2012, to February 16, 2013~\citep{yang2014modeling}.
The dataset includes 227,428 check-ins, each annotated with a timestamp, GPS coordinates, and detailed venue categories.
\textcolor{black}{We also replicate our analysis using a Foursquare dataset from Tokyo. 
Since the results closely align with those from New York City, we present the NYC results in the main text and report the Tokyo findings in Appendix~\ref{appendix:Tokyo}.}

For consistency, we assign a single category to each venue, selecting the second category level as it provides a balanced granularity between the broader first category and the highly specific third category.\footnote{\url{https://observablehq.com/d/94b009d907d7c023}}.
For example, a venue with the third category \emph{Italian Restaurant} is classified under the second category \emph{Restaurant} and the first category \emph{Dining and Drinking}.

We apply several preprocessing steps on the raw data. 
Specifically, we remove check-ins to categories representing familiar locations (e.g. \emph{Office, Home, Meeting Room}), modes of transportation (e.g. \emph{Train, Ferry, Road}), and venues labelled with \emph{Unknown} categories. 
Appendix~\ref{appendix:data} provides the list of excluded categories.

After preprocessing, the New York City dataset contains $166,306$ check-ins from $1,083$ unique users to $23,459$ unique venues that span $159$ distinct categories. On average, each category includes approximately 172 venues. 
In our simulations, we set $T_{\text{train}} = 210$ days and $T_{\text{max}} = 304$ days.

\subsection{Parameter settings}
\label{sec:parameters}
Our simulation framework relies on a set of parameters that govern the behavioural rules of the agents involved. Table \ref{tab:parameters} summarises the symbols used, their corresponding meanings, and the values adopted throughout the simulations. 
The framework requires only three parameters, aligning with our goal of assessing the effect of recommendation systems on urban dynamics with minimal changes to all other factors, thus isolating the impact of algorithmic intervention.

\begin{table}[htb!]
    \centering
    \caption{Simulation parameters adopted in our framework.}
    \def\arraystretch{1.2}
    \large
    \begin{tabular}{c|c|c}
    
       $\eta$ & adoption rate & $\{0.0, 0.2, 0.4, 0.6, 0.8, 1.0\}$ \\
       $\Delta$ &  retraining frequency & $7$ days \\
       $p$  & exploration probability & $0.6\times \lvert V \rvert^{-0.21}$ \citep{song2010modelling}\\
    \end{tabular}
    \label{tab:parameters}
\end{table}

The adoption rate $\eta$ is a key parameter in our framework, representing the probability that an individual follows the suggestion provided by the recommender system.
To systematically investigate the effects of recommendation systems, we vary $\eta$ across a predefined grid of values ranging from 0 to 1, with a step size of 0.2.

The parameter $\Delta$ reflects how frequently the recommendation system is updated (retrained) to account for the evolving nature of user preferences. 
We set $\Delta = 7$ days, meaning the recommender is retrained weekly to follow the dynamics introduced by the human-AI feedback loop. {\color{black}Results for $\Delta = 1$ day are similar; we provide them in Appendix~\ref{appendix:generalization}.}

When a human agent does not rely on the recommender system, it follows the EPR model \citep{song2010modelling}. 
Empirical analyses in \citet{song2010modelling} confirm the model's ability to reproduce the scaling properties of human mobility, capturing the balance between returning to familiar locations and exploring new ones. 
This dual behavioural perspective has been validated on real-world data, where the probability of exploring a new venue is given by $p=\rho\times \lvert V \rvert^{-\gamma}$, with $\lvert V \rvert$ denoting the number of unique venues. 
Following the empirical estimates of $\rho$ and $\gamma$ reported by \citet{song2010modelling}, we set their values to $0.6$ and $0.21$, respectively.

%% file: experiments.tex
\begin{figure}[htb!]
    \centering
    \includegraphics[width=0.95\linewidth]{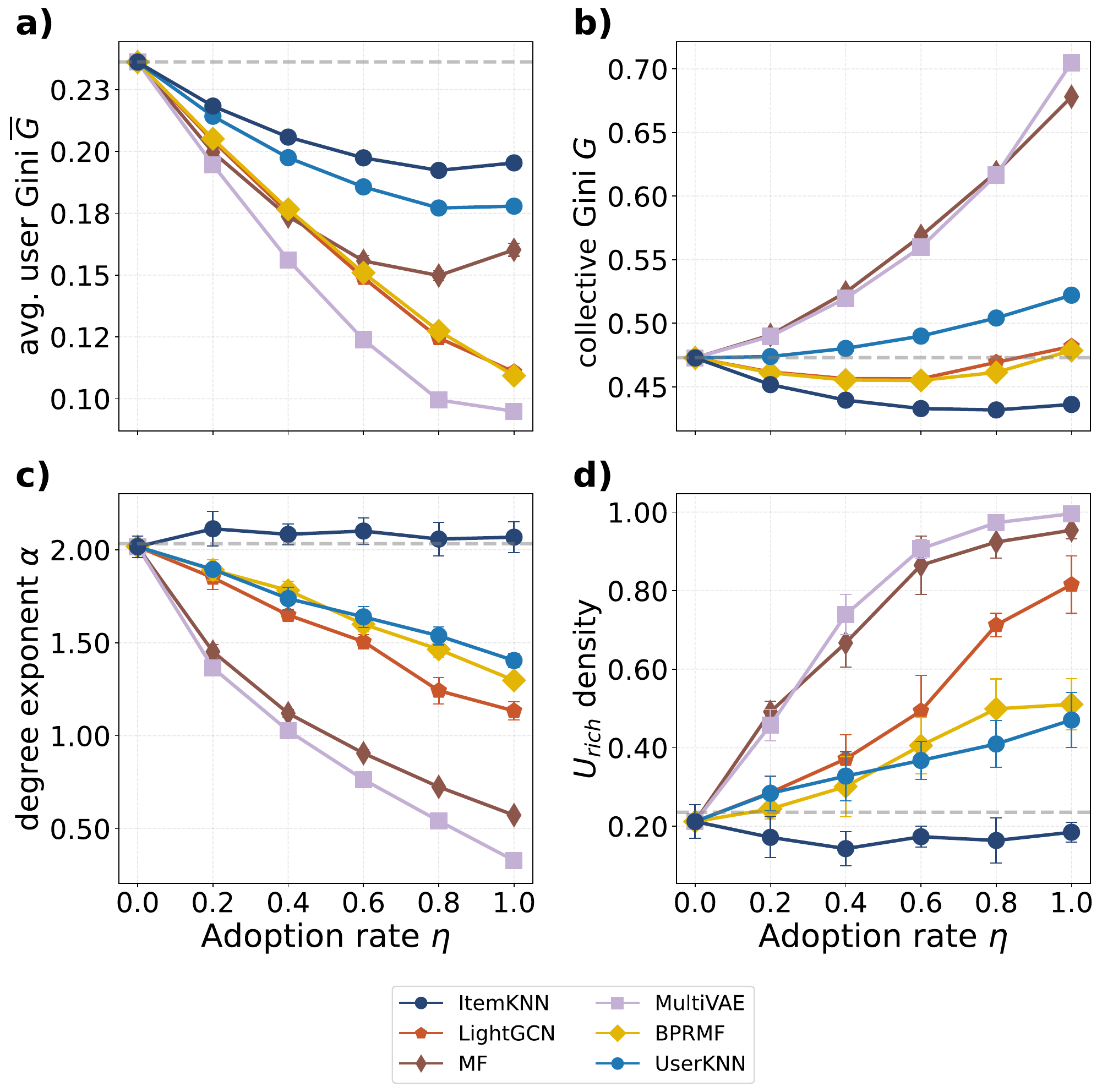}
\caption{\textbf{Effects of recommender system adoption $\eta$.}
(a) Average individual Gini coefficient $\overline{G}$;
(b) Gini coefficient of the distribution of visits across venues $G$;
(c) Absolute
value of the slope of the co-location network's degree distribution $\alpha$;
(d) Internal density of the rich club (top 15 nodes by degree) in the co-location network.
Each curve represents a recommender system; each point represents the average over five simulation runs; \textcolor{black}{the error bars represent the standard deviations.}
}
    \label{fig:comparison}
\end{figure}
 
Figure~\ref{fig:comparison} compares the effects of the feedback loop as a function of the adoption rate $\eta$. 
All values are averaged over five independent simulation runs.
Figure~\ref{fig:comparison}a and Figure \ref{fig:comparison}b report the average individual Gini coefficient $\overline{G}$ and the collective Gini coefficient $G$ of venue visits, respectively. 
We remind that the former captures how evenly each individual distributes their visits across venues, while the latter reflects overall disparities in venue popularity.

For the sake of brevity, we do not report results for PGN in the main text, as they closely mirror the patterns observed for MF and MultiVAE. 
A detailed comparison that includes PGN is provided in Appendix~\ref{app:pgn}.

The feedback loop consistently reduces $\overline{G}$ across all simulations, regardless of the specific recommender system embedded within it. Compared to the baseline with no algorithmic influence ($\eta = 0$, dashed line at $\overline{G} \approx 0.2$ in Figure~\ref{fig:comparison}a), the activation of the feedback loop leads to lower values of $\overline{G}$, with the effect intensifying as the adoption rate $\eta$ increases. While the feedback loop involving ItemKNN and UserKNN results in modest reductions, those involving deep learning-based models -- BPRMF, LightGCN, and MultiVAE -- produce significantly stronger decreases. Notably, for $\eta = 1$ (full reliance on the recommender system), the MultiVAE-based feedback loop reduces $\overline{G}$ to approximately 0.09, a relative decrease of about 60\% compared to the baseline. This result suggests that, when algorithmic influence is strong, individuals tend to diversify their mobility patterns, visiting a broader and more varied set of venues rather than repeatedly returning to the same few.

\begin{tcolorbox}[colback=gray!10, colframe=black!20, title=Key Result 1]
The feedback loop involving recommender systems boosts individual diversity, as reflected by a consistent drop in the average individual Gini coefficient.
\end{tcolorbox}

However, this increase in individual diversity does not necessarily translate into more diverse collective behaviour.
The feedback loop leads to an increase in the collective Gini coefficient $G$ with rising $\eta$ for some recommender systems (see Figure~\ref{fig:comparison}b).
UserKNN, MF, and MultiVAE all result in a sharp increase in $G$ as adoption grows.
In particular, for $\eta = 1$, the feedback loop with MultiVAE reaches $G = 0.7$, an increase of approximately 47\% compared to $\eta = 0$, indicating a more uneven distribution of visits across venues.
Deep learning–based models such as LightGCN and BPRMF contribute to slightly reducing inequality, except in full-adoption scenarios.
ItemKNN is the only model that slightly but consistently reduces global inequality compared to the baseline, reaching $G \approx 0.44$, a relative reduction of about 7\% compared to $\eta = 0$.
These results suggest that a systematic increase in individual diversity may be accompanied by a simultaneous concentration of visits on a smaller subset of venues.

\begin{tcolorbox}[colback=gray!10, colframe=black!20, title=Key Result 2]
More individual diversity does not imply more collective diversity: for some recommender systems, the feedback loop concentrates visits on a small set of popular venues.
\end{tcolorbox}

Analysis of the co-location network reveals that the feedback loop increases the absolute
value of the slope of the power-law degree distribution, $\alpha$, for all recommender systems except ItemKNN, with the effect particularly pronounced for MultiVAE and MF (see Figure~\ref{fig:comparison}c).
A lower slope $\alpha$ corresponds to a flatter distribution, indicating that co-location ties are more evenly distributed across individuals rather than concentrated among a few highly connected ones.
Thus, in most cases, the feedback loop reduces the skewness of the co-location structure, promoting broader interpersonal mixing.
In contrast, for ItemKNN, the slope $\alpha$ remains close to the baseline as the adoption rate $\eta$ increases, indicating little change in user co-presence patterns (squares in Figure~\ref{fig:comparison}c).
 
A similar pattern is observed in the rich-club structure $U_{\text{rich}}$, defined as the top 15 individuals with the highest degree in the co-location network. 
For all recommender systems except ItemKNN, the internal density of connections within this group -- the ratio of observed to possible links -- increases sharply with higher values of $\eta$, approaching nearly 1 for MultiVAE (see Figure~\ref{fig:comparison}d). 
This suggests that highly connected individuals increasingly tend to co-locate with one another. In contrast, for ItemKNN, the rich-club density remains close to the baseline level of approximately 0.2, indicating little change in elite clustering. 

\textcolor{black}{Results for Tokyo closely mirror those for New York City, with minor differences in rich-club density likely due to Tokyo’s higher initial network density and more compact geography (see Appendix~\ref{appendix:Tokyo}).}
\textcolor{black}{In addition, a robustness analysis using MultiVAE -- the recommender system showing the strongest effects -- confirms that our main findings are stable under alternative retraining configurations. 
Specifically, we tested both daily retraining and a sliding-window update scheme, observing no substantial differences in the results (see Appendix~\ref{appendix:generalization}).}

\begin{tcolorbox}[colback=gray!10, colframe=black!20, title=Key Result 3]
The feedback loop promotes broader mixing of co-locations, while rich-club individuals become nearly fully connected within the co-location network.
\end{tcolorbox}

\subsection*{Venue Perspective}
\begin{figure}[htb!]
    \centering
    \includegraphics[width=1.\linewidth]{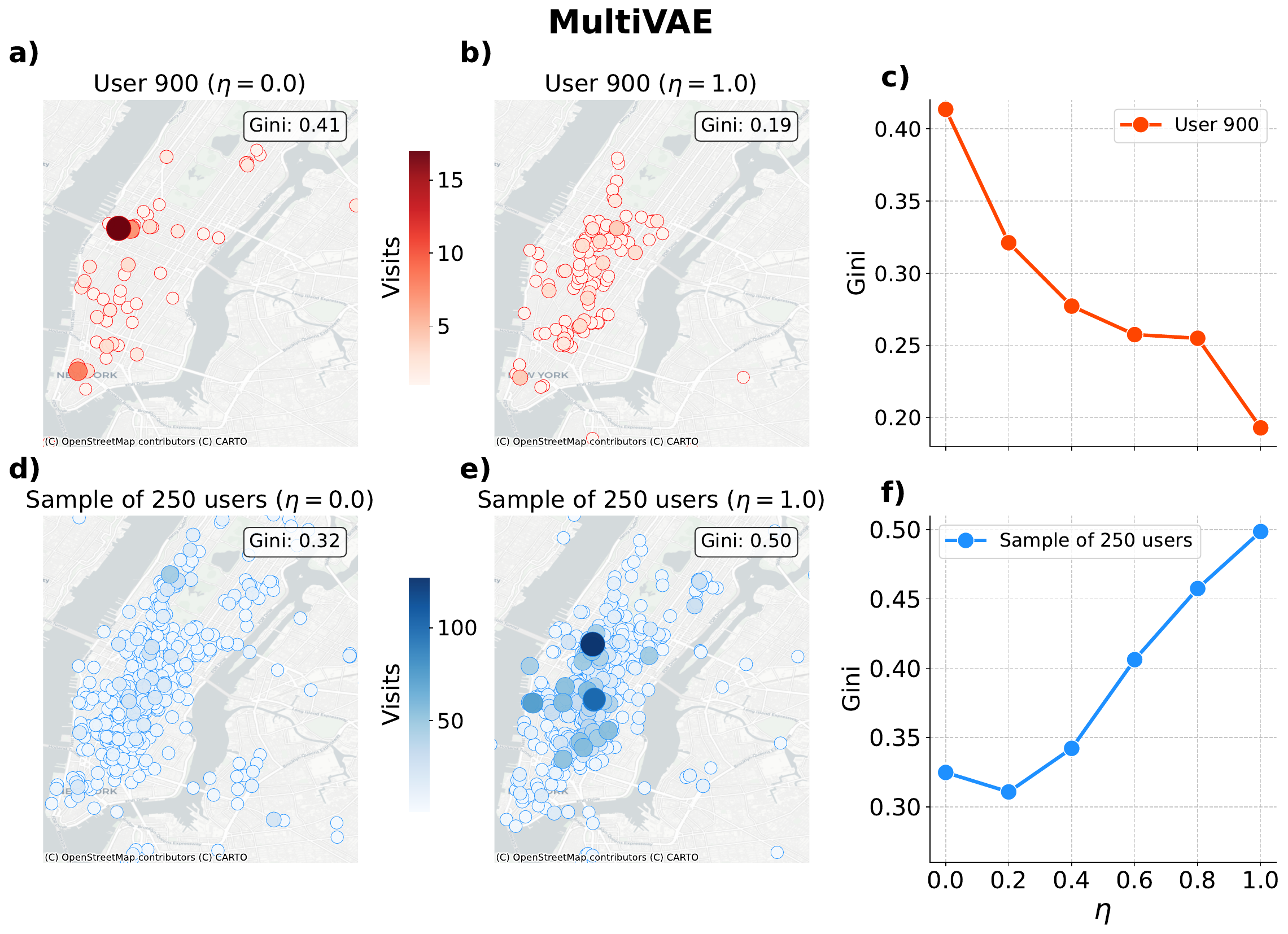}
\caption{\textbf{Venue visitation patterns under MultiVAE for null ($\eta = 0$) and full ($\eta = 1$) adoption.}
(a, b) Geographical distribution of visits by a representative user (User 900). 
Visit intensity is proportional to point size and color. 
When $\eta = 1$, visits become more evenly distributed across venues, with the individual Gini coefficient decreasing from $G_u = 0.41$ to $G_u = 0.19$.
(c) Individual Gini coefficient for User 900 as a function of $\eta$, showing a consistent decline.
(d, e) Aggregated venue visits for a sample of 250 users. 
When $\eta = 1$, visits are more concentrated in a few high-traffic venues, particularly in central Manhattan.
(f) Gini coefficient of venue visit distribution across the sample, rising from $G = 0.32$ to $G = 0.50$ with increasing $\eta$.}
    \label{fig:MultiVAE-spatial}
\end{figure}

\begin{figure}[htb!]
    \centering
    \includegraphics[width=1.\linewidth]{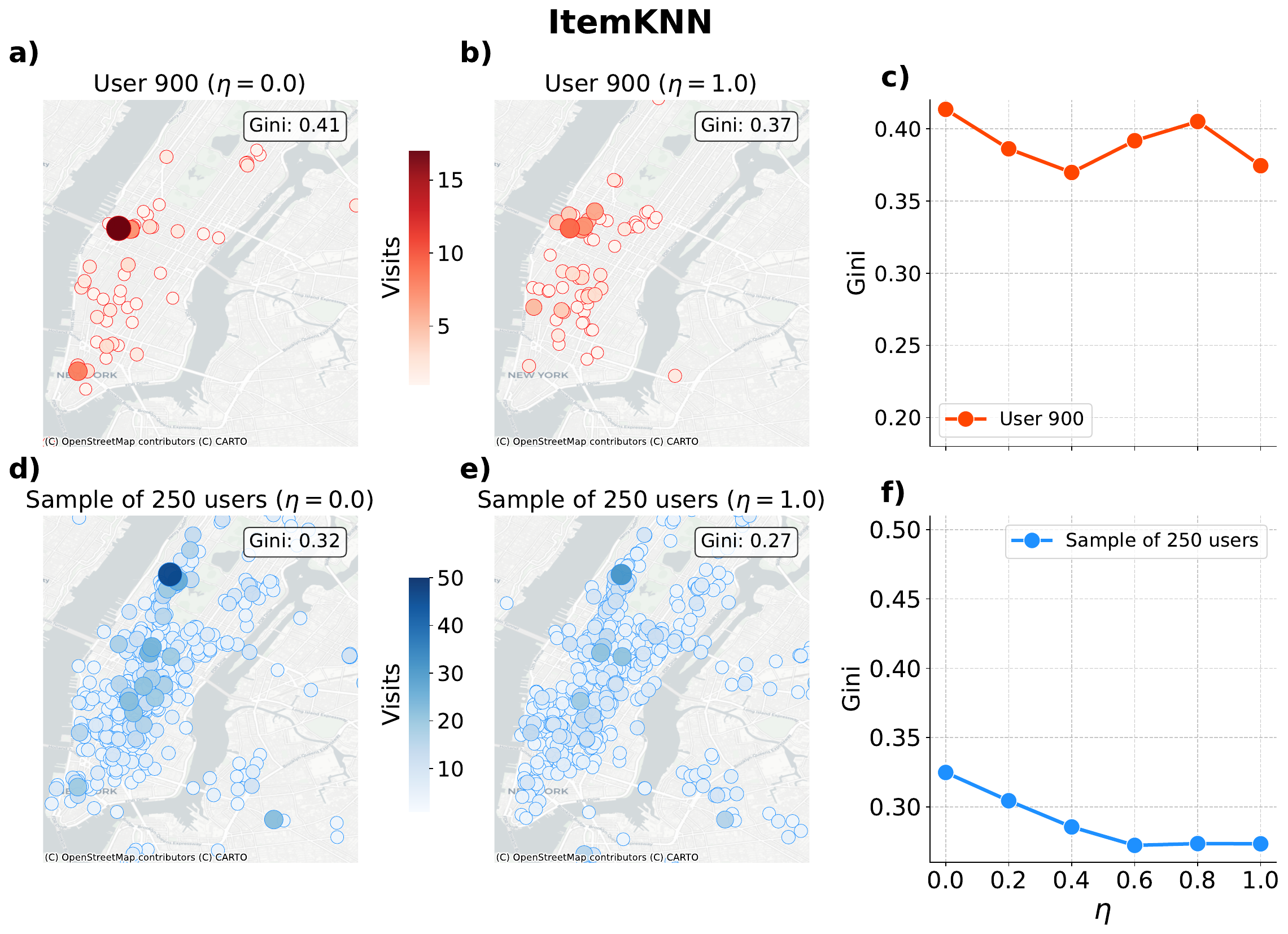}
    \caption{\textbf{Venue visitation patterns under ItemKNN for null ($\eta = 0$) and full ($\eta = 1$) adoption.}
(a, b) Geographical distribution of visits by a representative user (User 900). A modest shift toward a more balanced visitation pattern is observed, with the individual Gini decreasing from $G_u = 0.41$ to $G_u = 0.37$.
(c) Individual Gini coefficient for User 900 as a function of $\eta$, showing minor fluctuations.
(d, e) Aggregated venue visits for a sample of 250 users. Visit distributions remain fairly dispersed under both conditions.
(f) Gini coefficient of venue visit distribution across the sample, decreasing slightly from $G = 0.32$ to $G = 0.27$.}
    \label{fig:ItemKNN-spatial}
\end{figure}

Since the feedback loops involving MultiVAE and ItemKNN represent the two extreme cases across all the metrics discussed, we provide a more detailed analysis of the patterns that emerge from their integration within our framework.
The contrasting spatial effects of MultiVAE and ItemKNN are illustrated in Figures \ref{fig:MultiVAE-spatial} and \ref{fig:ItemKNN-spatial}, which show both individual and collective patterns under the extreme cases of null and full adoption ($\eta = 0$ and $\eta = 1$).

\paragraph{MultiVAE}
For MultiVAE, Figure \ref{fig:MultiVAE-spatial}a-b shows the spatial distribution of visits for a representative individual (User 900). 
Under the scenario of no recommendation ($\eta = 0$), visits are mostly concentrated on just two venues in western Midtown Manhattan (Figure \ref{fig:MultiVAE-spatial}a). 
Under full adoption ($\eta = 1$), the activity of User 900 becomes more spatially dispersed, with visits more evenly distributed throughout the city (Figure \ref{fig:MultiVAE-spatial}b). 
This shift is captured by the individual Gini coefficient, which drops from $G_u = 0.41$ to $G_u = 0.19$ -- a reduction of approximately 54\% -- with a consistent decline observed as the adoption rate $\eta$ increases (Figure \ref{fig:MultiVAE-spatial}c).

At the collective level, MultiVAE shows the opposite effect. Figure \ref{fig:MultiVAE-spatial}d-e shows the aggregated spatial distribution of visits for a random sample of 250 users. When $\eta = 0$, visits are relatively uniformly spread across the city, including peripheral neighbourhoods and outer boroughs, with $G=0.32$ (see Figure \ref{fig:MultiVAE-spatial}d). 
When $\eta = 1$, visits become concentrated in a small number of venues, especially in Midtown and Lower Manhattan. 
The Gini coefficient increases to $G=0.50$, marking a 56\% rise in venue popularity inequality (see Figure \ref{fig:MultiVAE-spatial}f). 

These results confirm the results shown in the previous section: while the feedback loop with MultiVAE promotes more balanced behaviour at the individual level, it drives convergence toward a narrow set of highly frequented locations at the collective level, reinforcing popularity inequality.
This pattern is typical of the feedback loops involving most other recommender systems we examined.

\paragraph{ItemKNN}
ItemKNN produces more stable spatial dynamics than MultiVAE (see Figure \ref{fig:ItemKNN-spatial}). 
For User 900, the shift from $\eta = 0$ to $\eta = 1$ results in a slightly more balanced, yet still unequal, visitation pattern (see Figures \ref{fig:ItemKNN-spatial}a-b). 
The individual Gini coefficient decreases from $G_u=0.41$ to $G_u=0.37$, a reduction of approximately 9.8\%, with only minor fluctuations across $\eta$ (Figure \ref{fig:ItemKNN-spatial}c).
At the collective level, ItemKNN contributes to a small improvement in spatial balance. Figure \ref{fig:ItemKNN-spatial}d-e shows that the city-wide distribution of visits remains fairly dispersed across both conditions. 
The Gini coefficient decreases from $G=0.32$ to $G=0.27$ under full adoption (Figure \ref{fig:ItemKNN-spatial}f confirms this downward trend). 
Unlike MultiVAE, the feedback loop involving ItemKNN does not lead to spatial convergence around a narrow set of venues.
Instead, it preserves heterogeneity in mobility patterns while slightly enhancing the overall equity in how urban space is visited.

\subsection*{Individual Perspective}
\begin{figure}[hbt!]
\centering
\begin{subfigure}{\textwidth}  
  \centering
  \includegraphics[width=\textwidth]{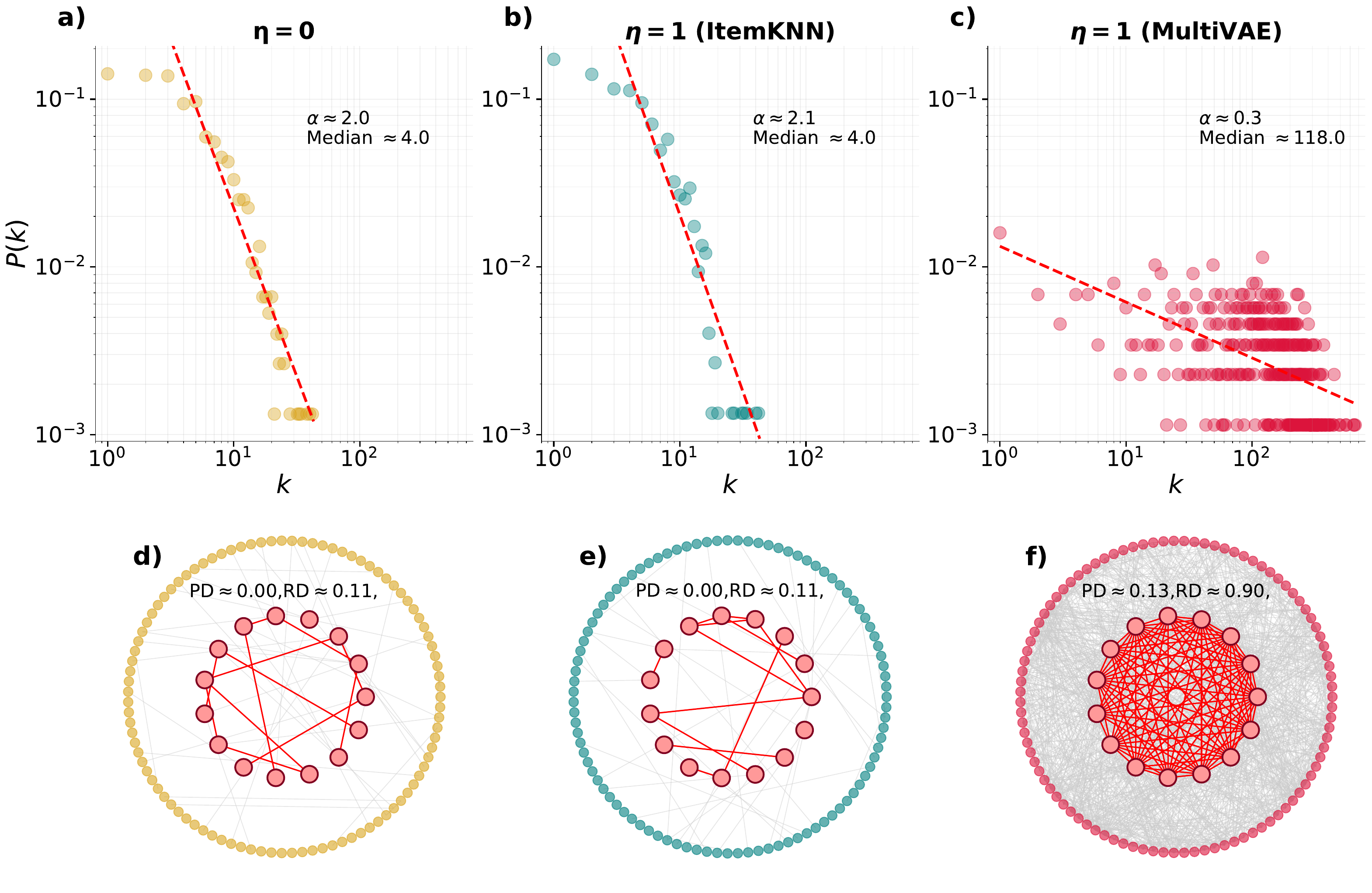}  
  \label{fig-vs1}
\end{subfigure}

\vspace{1em}  

\begin{subfigure}{.9\textwidth}  
  \centering
  \includegraphics[width=.8\textwidth]{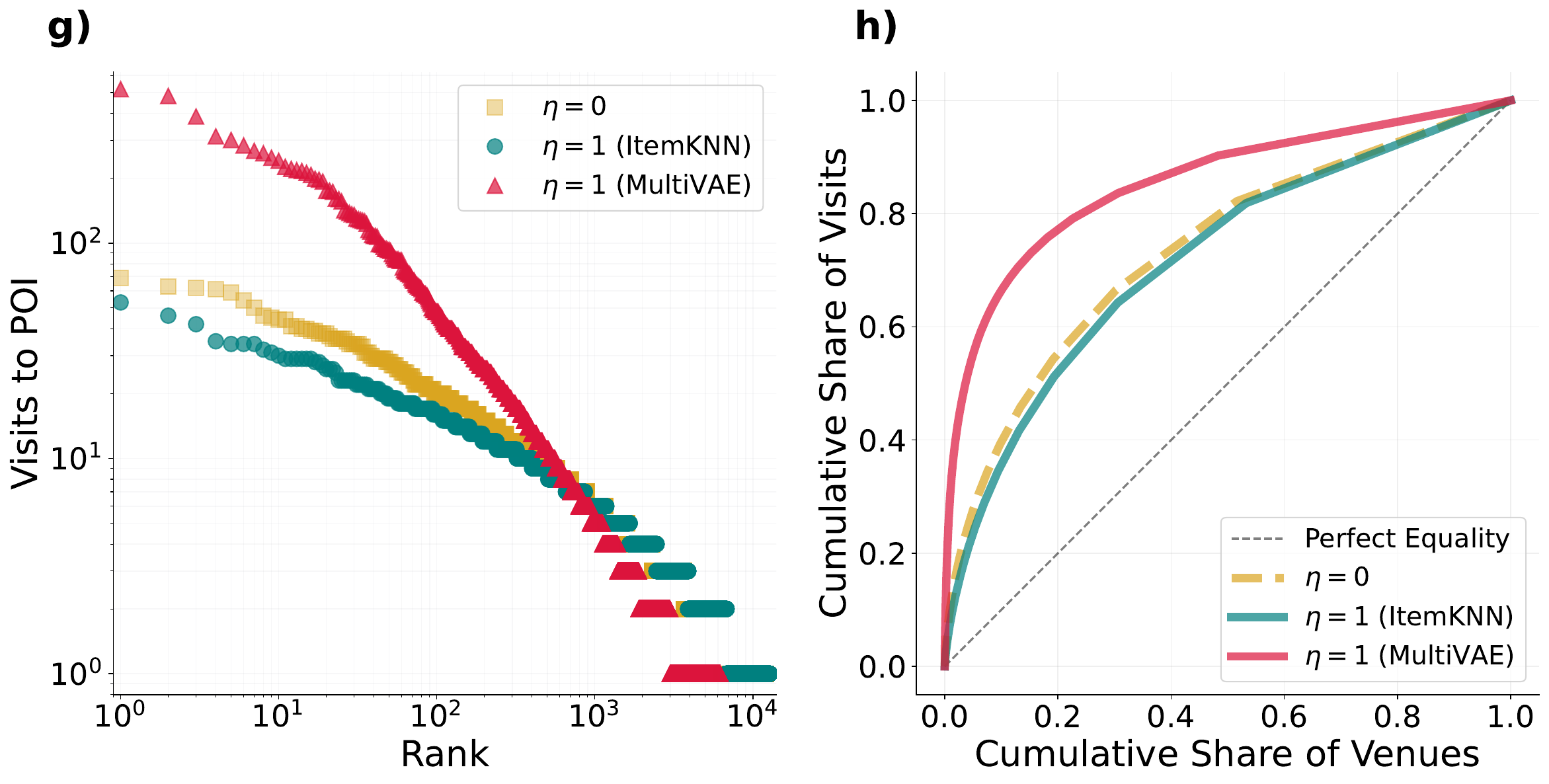}  
  \label{fig-vs2}  
\end{subfigure}
\caption{
\textbf{Effects of the feedback loop under null ($\eta = 0$) and full ($\eta = 1$) adoption.}
(a-c) Degree distribution of the co-location network, for null adoption, full adoption with ItemKNN and full adoption with MultiVAE. MultiVAE significantly flattens the distribution and increases median degree, while ItemKNN maintains the original network structure.
(d-f) Visualization of the co-location network (sampled), highlighting the rich-club structure (red nodes) and peripheral nodes (circular layout) for null adoption, full adoption with ItemKNN and full adoption with MultiVAE. Rich-club density increases dramatically for MultiVAE, while remaining constant for ItemKNN. 
(g-h) Rank-size distribution and Lorenz curves of venue visits: under $\eta {=} 1$ for MultiVAE, top-ranked venues receive disproportionately more visits.
}
\label{fig:social_analysis}
\end{figure}

\paragraph{MultiVAE}
The feedback loop involving MultiVAE substantially reshapes the structure of urban co-presence (see Figure \ref{fig:social_analysis}). 
We remind that, in the co-location network, two individuals are linked if they visit the same venue within the same epoch of the simulation.
Figures \ref{fig:social_analysis}a,c show the degree distributions of the co-location network under the two extreme cases of null and full adoption ($\eta = 0$ and $\eta = 1$). 
Under the autonomous choice scenario ($\eta=0$), the degree distribution follows a steeper power-law decay, with a median degree of 4 and the absolute value of the slope $\alpha \approx 2.0$ (Figure \ref{fig:social_analysis}a). 
When users fully adopt MultiVAE ($\eta = 1$), the degree distribution flattens significantly, with a heavier tail and a substantial increase in median degree (Figure \ref{fig:social_analysis}c). 
Notably, the number of nodes in the network also increases, which implies broader participation in the social layer under algorithmic influence.

Figure \ref{fig:social_analysis}d,f offers a visual snapshot of the co-location network structure under the null and full adoption scenarios for MultiVAE. 
The inner red nodes denote the rich club and the peripheral nodes lie along the outer circle. 
When $\eta = 0$, the peripheral subnetwork is sparse (Peripheral Density, PD is almost 0) and the rich club is weakly connected (Rich-club Density, $RD \approx 0.11$). When $\eta = 1$ with MultiVAE, the structure shifts dramatically: peripheral nodes become more interconnected ($PD \approx 0.13$) and the rich club much more densely connected ($RD \approx 0.90$), forming a dense social core.

These structural shifts in co-location patterns are tightly connected to changes in venue popularity. Figure \ref{fig:social_analysis}g reports the rank-size distribution of venue visits. 
With full adoption of MultiVAE, the head of the distribution grows significantly: a few venues receive disproportionately high attention, reflecting an increase in spatial centralisation. 
This dynamic reinforces not only high-degree users but also high-frequency venues, suggesting that popular places become even more dominant.
This pattern is further confirmed in Figure \ref{fig:social_analysis}h, which reports the Lorenz curves of visit distributions under different scenarios. The curve for MultiVAE with $\eta = 1$ deviates more strongly from the equality line, indicating higher overall inequality in venue visitation, despite the increased uniformity in individual user behaviour and co-location degree.

\paragraph{ItemKNN}
In contrast to MultiVAE, the feedback loop involving ItemKNN has a minimal effect on the structure of the co-location network and the distribution of visits across venues. 
Figure \ref{fig:social_analysis}a,b show the degree distributions of the co-location network under $\eta = 0$ and $\eta = 1$ with ItemKNN, respectively. 
Both distributions follow similar power-law trends, with nearly identical slopes ($\alpha \approx 2.0$ and $\alpha \approx 2.1$) and a stable median degree of 4. 
This indicates that the feedback loop involving ItemKNN does not substantially alter the topology of social co-presence: individuals maintain similar numbers of co-located interactions regardless of whether they follow the algorithm's suggestions.

A structural view of the co-location network confirms this stability. 
Figure \ref{fig:social_analysis}d,e visualises the rich-club structure under null adoption and full adoption with ItemKNN. 
In both cases, the peripheral density remains at zero ($PD \approx 0.00$), and the rich-club density remains almost constant ($RD = 0.11$ and $RD \approx 0.11$ respectively). 
This suggests that the feedback loop with ItemKNN neither strengthens the elite core of highly connected individuals nor promotes wider cohesion among peripheral ones.

The implications from a venue perspective are shown in Figure \ref{fig:social_analysis}g-h. 
The rank-size distribution of venue visits (Figure \ref{fig:social_analysis}g) shows that ItemKNN largely preserves the pattern of venue visits across adoption levels, with only minor variation in the long tail.
The Lorenz curves (Figure \ref{fig:social_analysis}h) reveal a slight shift towards greater equality under full adoption of ItemKNN, as its curve moves marginally closer to the diagonal compared to the null adoption scenario. 
This indicates a modest reduction in the inequality of venue visits, although the general pattern remains similar.

Overall, the feedback loop involving ItemKNN preserves the existing structure of user interactions and venue usage. 
Unlike more sophisticated recommender systems, it avoids concentrating visits or reinforcing strong social cores, instead maintaining a relatively stable distribution of co-location and visits across the system.


\subsection{Understanding the individual-collective diversity trade-off}

Within the feedback loop, the recommender systems analysed reduce inequality in how individuals distribute their visits across venues, yet most recommenders simultaneously increase the inequality in how visits are distributed across venues at the collective level.
How can these two seemingly contradictory mechanisms coexist?

To address this question, we compare the exploratory visits of individuals during the simulation with those recorded in the historical training data ($D_{\text{train}}$). 
We model the interaction between individuals and venues as a bipartite network, where one set of nodes represents individuals, the other represents venues, and edges correspond to visits. 
We construct two bipartite networks: one based on the historical training data, and another based on exploratory visits simulated under full algorithmic adoption ($\eta=1$). In the training network, an edge connects an individual to a venue if it was visited during the training period. In the simulation network, an edge is added only if the individual visited a venue for the first time, i.e., a venue not in $D_{\text{train}}$.

We focus on MultiVAE, which exhibits the strongest impact on inequality patterns. 
We group venues into deciles based on their number of unique visitors in the training set. Each decile represents 10\% of venues, ranked from least (decile 1) to most popular (decile 10).

The degree of a venue in the bipartite network reflects how many users visited it. Since we model binary interactions, the degree is equivalent to the weighted degree. To normalise across the network, we compute the normalised weighted degree for each venue \(v\):

\begin{equation}
\hat{d}_v = \frac{d_v}{\sum_{v' \in V} d_{v'}}
\end{equation}

We then aggregate these values within each decile to compute the share of attention directed to that group:

\begin{equation}
\hat{D}_g = \sum_{v \in g} \hat{d}_v
\end{equation}

\noindent This measure, \(\hat{D}_g\), captures the fraction of all user–venue interactions involving venues in the group \(g\). 


\begin{figure}[t]
\centering
\begin{subfigure}{0.5\textwidth}
  \centering
  \includegraphics[width=\textwidth]{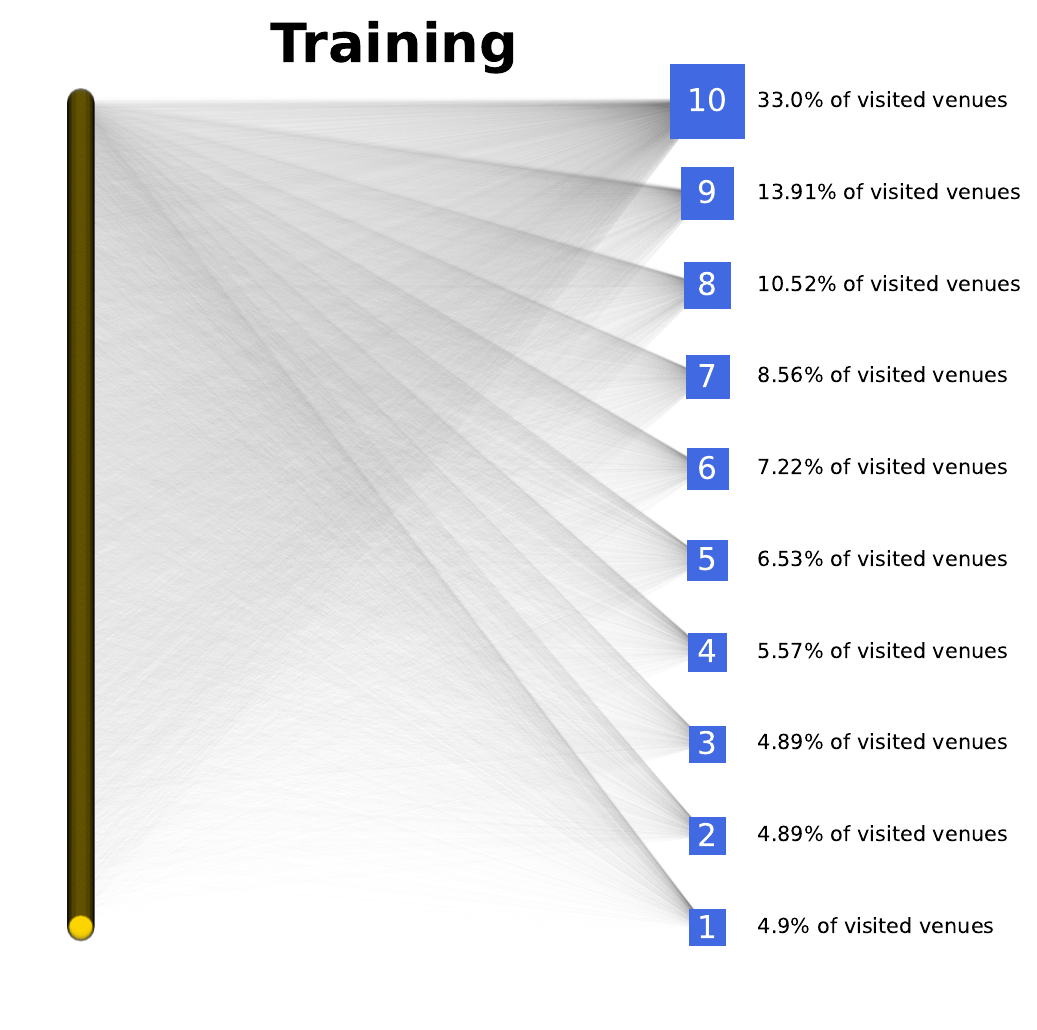}
  \caption{Training set network}
  \label{fig-bipartite-train}
\end{subfigure}%
\begin{subfigure}{0.5\textwidth}
  \centering
  \includegraphics[width=\textwidth]{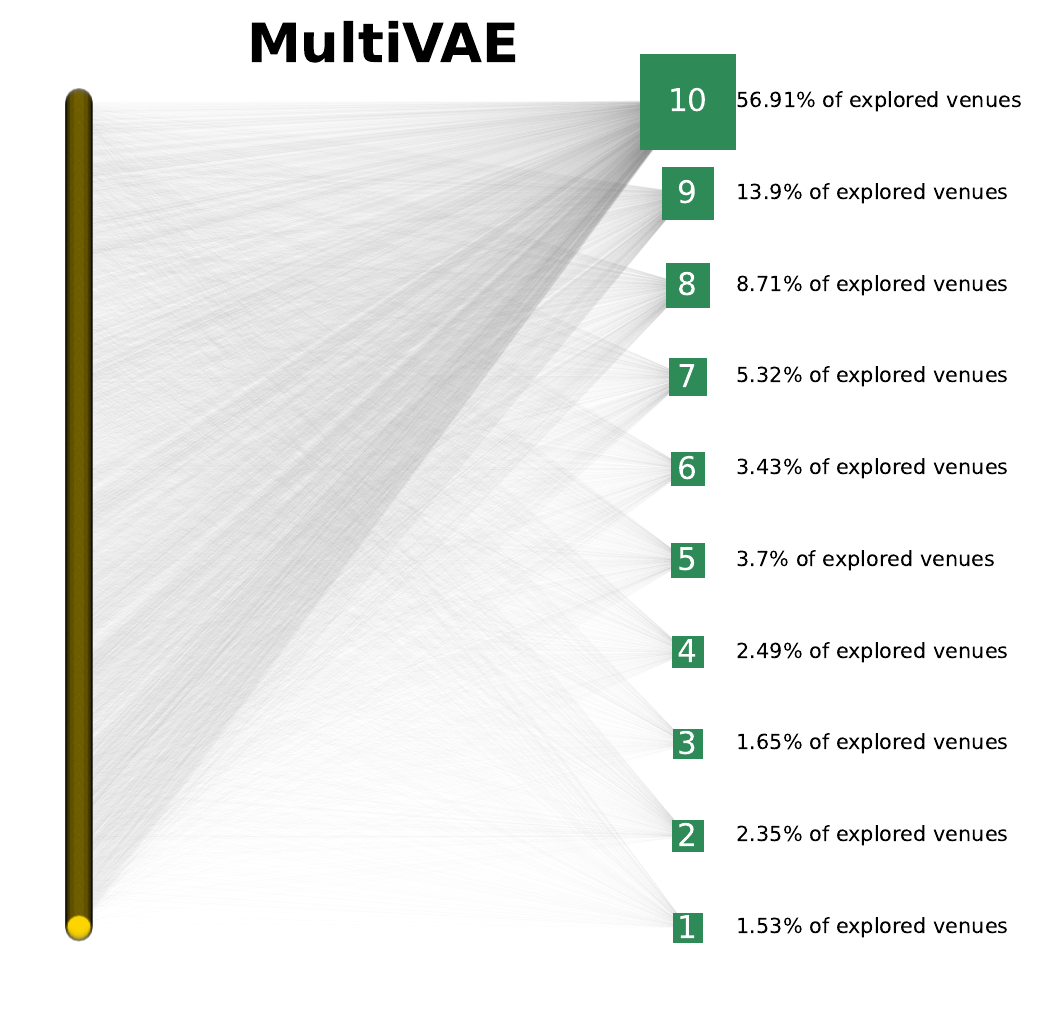}
  \caption{Exploration network (MultiVAE)}
  \label{fig-bipartite-exp}
\end{subfigure}
\caption{\textbf{User-venue bipartite networks.} In the bipartite network, individuals are on the left and venue deciles are on the right. 
Each decile contains 10\% of venues, ranked by their popularity in the training data. 
Percentages indicate each decile’s share of total user–venue interactions. In the bipartite network corresponding to the training data (a), an edge indicates a visit during the training period. 
In the bipartite network of the simulation data (b), an edge represents a simulated visit to a venue not seen during training (i.e., a novel exploration). Venues with fewer than 3 unique visitors are excluded.}
\label{fig-combined-networks}
\end{figure}

Figure~\ref{fig-bipartite-train} shows that while the most popular venues (decile 10) already attract the largest share of attention (33\%) in the training data, the other deciles also receive a significant portion of visits. 
After the simulation, under the influence of MultiVAE, attention becomes far more concentrated (see Figure~\ref{fig-bipartite-exp}): decile 10 captures 56.91\% of all interactions in the exploration phase, an increase of nearly 24\% compared to the training data. 
This comes at the expense of nearly every other decile, whose shares decline.

\begin{figure}[t]
\centering
\includegraphics[width=0.5\textwidth]{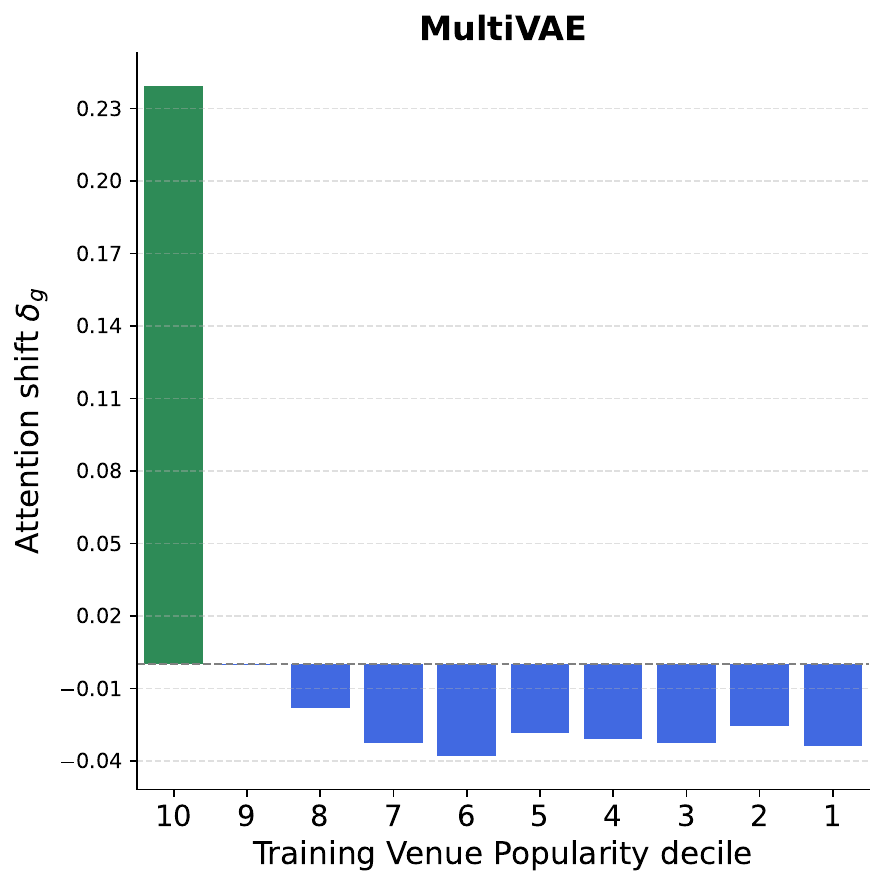}
\caption{\textbf{Change in normalized weighted degree across venue popularity deciles (1–10).} 
Bars show the difference in share of interactions between exploration and training under MultiVAE.}
\label{fig-decile-diff}
\end{figure}

To characterise this shift better, we compute the change in attention share for each decile between the training data and simulation data:
$
\delta_g = \hat{D}_g^{\text{(simulation)}} - \hat{D}_g^{\text{(train)}}
$.
As shown in Figure~\ref{fig-decile-diff}, the shift is stark: decile 10 gains over 20 percentage points in attention, while all other deciles, especially the least popular ones, loose ground. 

The analysis shows that the new venues visited by individuals, those driving the increase in individual diversity, are often the same across the population, typically the most popular ones. 
This reflects a rich-get-richer dynamic: while individual-level diversity improves, it does so by directing many individuals toward a narrow set of highly recommended venues. 
As a result, the system becomes more centralised, amplifying popularity bias and revealing a fundamental tension: recommendation systems can enhance fairness at the individual level while undermining it collectively.
Appendix~\ref{appendix:similarity} offers a complementary perspective on this dichotomy, showing that user similarity increases with the adoption rate~$\eta$ for complex, representation-based models such as MultiVAE, while remaining stable for ItemKNN.

\begin{tcolorbox}[colback=gray!10, colframe=black!20, title=Key Result 4]
Individual diversity rises mostly through visits to the same popular venues.
\end{tcolorbox}

%% file: conclusions.tex
In this article, we modelled the human–AI feedback loop in the context of next-venue recommendation to simulate its impact on the diversity of visits at the individual and collective level, and on the structure of individuals' co-location networks. 

Our study has some limitations. 
First, our analysis of the feedback loop is based on simulation rather than empirical observation, due to the lack of access to real-time platform data such as recommendation outputs and user interactions. 
As a result, our analysis should be intended as a what-if analysis to explore hypothetical scenarios rather than providing empirical evidence. 
Regulatory initiatives such as the Digital Services Act will empower researchers to run empirical experiments directly on online digital platforms, enabling rigorous assessments of how recommender systems influence real-world urban dynamics. 
In the meantime, our work offers a set of testable conjetures -- the key results in the grey boxes -- that can guide future empirical investigations on these platforms.

{\color{black}Second, we model individual decision-making using simplified probabilistic rules, which do not fully capture the complexity of human behaviour. 
For example, we assume the same probability of adopting the recommender system across all individuals, without explicitly accounting for individual differences in receptiveness to recommendations. 
Our assumption of the same adoption probability across all individuals is supported by the findings of \cite{cinus2022effect}, which suggest that the qualitative effects of recommender systems remain robust even when individual susceptibility to recommendations is heterogeneous.
Nonetheless, incorporating user-level heterogeneity more explicitly, or designing simulations with persistent exposure conditions, would provide a valuable extension to better understand the nuanced interplay between algorithmic influence and venue visitation patterns.}
We plan to address the limitations discussed above in future works.

Our study opens several directions for future research. 
One promising avenue is to enrich the autonomous decision-making mechanism by integrating more sophisticated mechanistic models of human mobility, capturing factors such as temporal routines \citep{jiang2016timegeo, pappalardo2018data}, social interactions \citep{toole2015coupling, cornacchia2021mechanistic}, and spatial scales \citep{alessandretti2020scales}. 
{\color{black} In this regard, an interesting direction is to model the decision-making process as a two-step sequence: first selecting an activity category (e.g., dining, shopping, entertainment), and then choosing a specific venue within that category. 
This would allow us to capture how recommender systems influence not just where individuals go, but also what types of activities they choose to engage in. 
Another possible improvement would be to account for commercial mechanisms typically implemented in online platforms, such as venue sponsorship or monetized rankings.
These mechanisms can alter the visibility of certain venues and may therefore influence the dynamics observed in this study.}



\textcolor{black}{Another interesting direction is the mitigation of the observed biases in next-venue recommendation. 
Model ensembling offers a promising approach in this regard. 
While our study does not explicitly evaluate ensemble methods, it includes PGN \citep{sanchez2022travelers}, an ensemble model that integrates collaborative, geographical-proximity, and popularity signals. PGN displays behavioural patterns similar to deep learning models like MultiVAE, suggesting that even architectures integrating diverse signals may still reproduce certain biases. This underscores the need for future research to systematically assess how ensembling and signal integration impact fairness and diversity in human travel patterns.}

Concerning the field of human mobility, a direction for future work is to investigate how the human-AI feedback loop may influence the well-established statistical laws that characterise human movement by guiding individuals towards certain venues over others. 
It would be also important to explore whether the feedback loop reinforces or mitigates the socio-economic and demographic segregation that often emerges in urban visitation patterns \citep{moro2021mobility, barbosa2021uncovering, gambetta2023mobility}, potentially amplifying inequalities or, conversely, acting as tools for urban inclusion.
By shaping venue popularity, next-venue recommender systems can also affect property values and the perceived attractiveness of neighbourhoods, potentially driving complex urban processes such as gentrification and overtourism~\citep{mauro2025dynamic}.

In conclusion, our study represents the first attempt to study how the human-AI feedback loop reshapes human mobility patterns. 
Our novel computational framework opens new avenues for studying the coevolution between algorithmic decision support and individual choice in the city, contributing to a deeper understanding of how digital platforms shape, and are shaped by, human mobility patterns.

%% file: appendix.tex
%

\newpage
\section{Training process and \textcolor{black}{hyperparameters}}
\label{appendix:recsys}
UserKNN, ItemKNN, and PGN are trained using $k = 10$ nearest neighbors.
MF is implemented with 32 latent factors.
BPRMF and LightGCN are optimized using the pairwise BPR loss with a batch size of 16 and an L2 regularization term of 0.0001.
MultiVAE is trained with cross-entropy loss combined with a KL divergence term (weighted by a coefficient of 0.2) in a full-batch setting.

During the training process, we track the training loss at the end of each epoch and apply early stopping if the loss remains stable for five consecutive epochs. The total number of epochs is capped at 500. 
We use the Adam optimiser with a learning rate of 0.001.
{\color{black}
The hyperparameter values we use in our experiments are reported in Table \ref{tab:all_parameters}. }

As a sanity check, we evaluate each trained model on $D_{\text{train}}$ and $D_{\text{post}}$. For each user $u \in U$, we generate a top-K list of recommended venues and measure the HitRate and mean Reciprocal Rank (mRR) in predicting $u$’s interactions in $D_{\text{post}}$.  We set $K=20$ and report average HitRate and mRR across all users in Table \ref{tab:performance}.

\begin{table*}[ht]
  \centering
  \caption{Performance Metrics for Various Recommender Models on New York City dataset}
  \label{tab:performance}
  \begin{tabular}{lcc}
    \toprule
    Model      & HitRate@20 & mRR@20 \\
    \midrule
    UserKNN    & 0.1726     & 0.0576  \\
    ItemKNN    & 0.1703     & 0.0377  \\
    MF         & 0.1870     & 0.0474  \\
    BPRMF      & 0.2261     & 0.0611  \\
    MultiVAE   & 0.1898     & 0.0656  \\
    LightGCN   & 0.2774     & 0.0838  \\
    PGN        & 0.2156     & 0.0704  \\
    \bottomrule
  \end{tabular}
\end{table*}

The performance of the recommender systems aligns with findings reported in the literature \citep{sanchez2022travelers}: PGN outperforms BPRMF, which in turn exceeds UserKNN, followed by ItemKNN.

\begin{table}[htb!]
    \centering
    \caption{Complete set of simulation parameters adopted in our framework.}
    \def\arraystretch{1.2}
    \large
    \begin{tabular}{c|c|c}
       $\eta$ & adoption rate & $\{0.0, 0.2, 0.4, 0.6, 0.8, 1.0\}$ \\
       $\Delta$ &  retraining frequency & \{$1$ days, $7$ days\} \\
       $p$  & exploration probability & $0.6\times \lvert V \rvert^{-0.21}$ \citep{song2010modelling} \\
       {}  & training approach & \{incremental, sliding window\} \\
       {}  & optimizer & Adam \\
       {} & learning rate & 0.001 \\
       {} & epochs & 500 \\
       {} & early-stopping patience & 5 \\
       {} & batch size & \{16, 32\} \\
       {} & nr of nearest neighbors & 10 \\
       {} & KL divergence coeff (MultiVAE) & 0.2 \\
    \end{tabular}
    \label{tab:all_parameters}
\end{table}

{\color{black}
\section{Tokyo Dataset}
\label{appendix:Tokyo}
The Foursquare Tokyo dataset contains 573,703 check-ins from 2,293 users across 61,858 venues. 
This is approximately twice the size of the New York City dataset, while the geographic area covered by Tokyo --measured as the bounding box of all venues -- is roughly half that of New York City. 

To ensure a fair comparison between the two datasets, we randomly sample 1,083 users -- the same number as in the raw New York City dataset -- and apply the same preprocessing pipeline to both datasets.
The resulting Tokyo dataset contains 141,968 check-ins to 34,314 venues by 1,083 users.

\begin{figure}[h]
\centering
\includegraphics[width=0.85\linewidth]{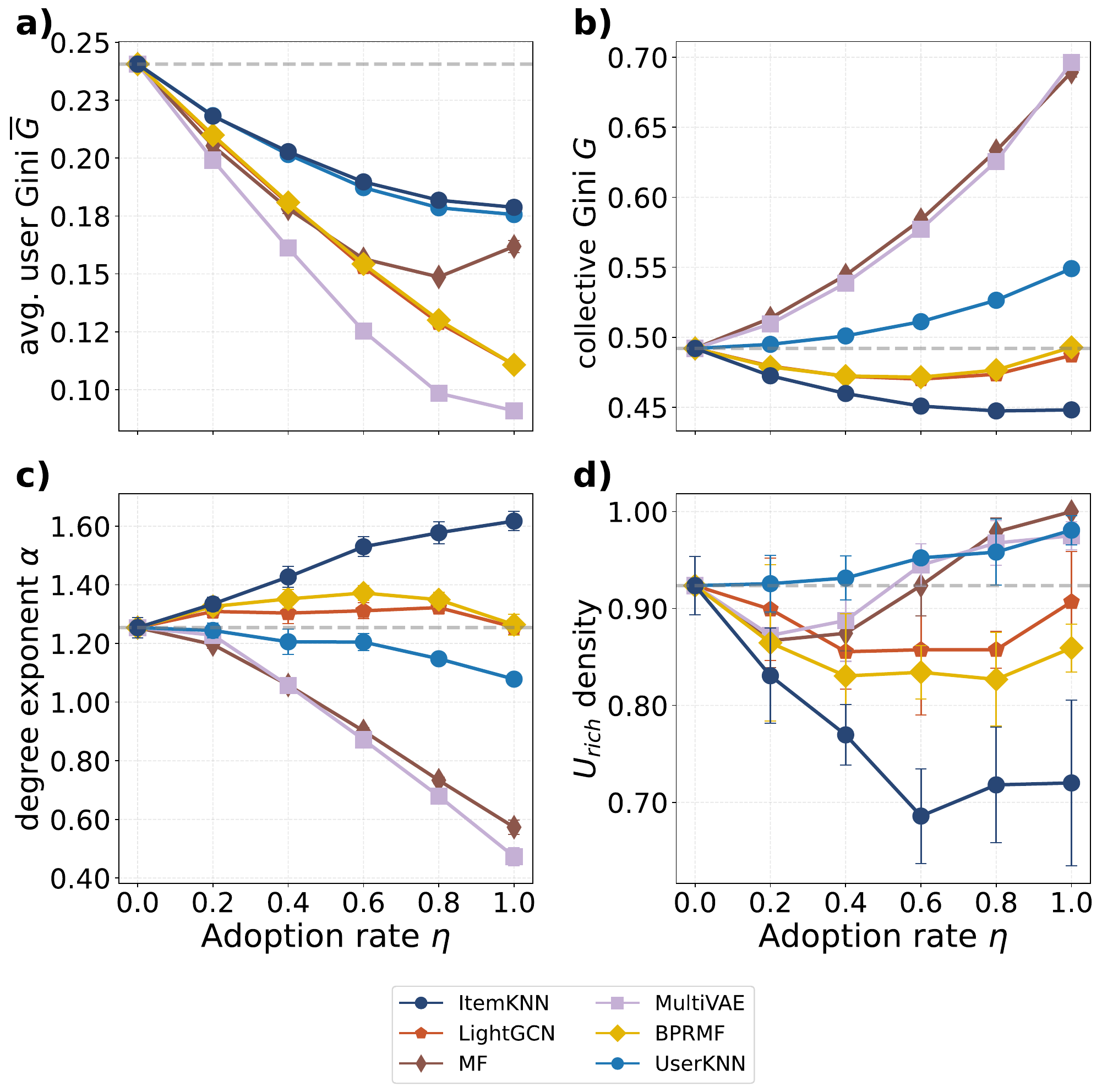}
\caption{Effects of different recommender systems on mobility and network-level indicators in Tokyo, as a function of the adoption rate $\eta$. Each point represents an average over five simulation runs. (a) Average individual Gini coefficient $\overline{G}$; (b) Collective Gini coefficient $G$; (c) Degree exponent $\alpha$ of the co-location network; (d) Rich-club density $U_{\text{rich}}$.}
\label{fig:tokyo_comparison}
\end{figure}

Figure~\ref{fig:tokyo_comparison} shows the impact of the feedback loop in Tokyo as a function of the adoption rate $\eta$, averaged over five simulation runs. 
The results closely mirror those observed for New York City, supporting the robustness of the main findings across different urban contexts.

Figure~\ref{fig:tokyo_comparison}a shows that all models reduce the average individual Gini coefficient $\overline{G}$ with increasing $\eta$, indicating that recommenders consistently promote more diverse individual behavior. MultiVAE again achieves the largest reduction, reaching values below $\overline{G}=0.1$ at full adoption, while the impact of KNN-based models is limited.
Figure~\ref{fig:tokyo_comparison}b reports the collective Gini coefficient $G$, which increases for most models as adoption rises. MultiVAE and MF lead to the strongest increases, similarly to what was observed in New York City, reaching values close to $G=0.7$. ItemKNN remains the only recommender system that consistently reduces collective inequality, similarly to what observed for New York City.

The degree exponent $\alpha$ of the co-location network is shown in Figure~\ref{fig:tokyo_comparison}c. 
As observed in New York City, most recommender systems flatten the degree distribution by lowering $\alpha$, promoting broader co-location across users. The effect is the strongest for MF, MultiVAE, and LightGCN, while ItemKNN and UserKNN exhibit more stable values near the baseline.
Figure~\ref{fig:tokyo_comparison}d presents the rich-club density $U_{\text{rich}}$, defined as the internal density among the 15 most connected individuals. At first glance, results differ slightly from those in New York City. 
In Tokyo, the initial rich-club density is already high (around 0.9) at $\eta = 0$, compared to about 0.2 in New York City. 
As a result, for most recommender systems, the feedback loop either maintains or slightly reduces $U_{\text{rich}}$, with the exception of MF, MultiVAE, and UserKNN, which preserve or further increase rich-club connectivity. These differences may be explained by the higher population density, more compact spatial distribution of venues, or contextual behavioral patterns specific to Tokyo.
Taken together, these results demonstrate that while some quantitative differences emerge -- particularly in the rich-club structure -- the main trends observed in New York City hold also in Tokyo. 
This reinforces the generalizability of our findings across different urban environments.

}

\section{Dataset cleaning procedure}
\label{appendix:data}
We excluded all ckeck-ins to venues with following categories: Train, Transport Hub, Transportation Service, Travel and Transportation, Boat or Ferry, Platform, Road, Island, River, Housing Development, Meeting Room, Conference Room, Office, Home (private), Apartment or Condo and Unknown.

\begin{figure}
    \centering
    \includegraphics[width=0.75\linewidth]{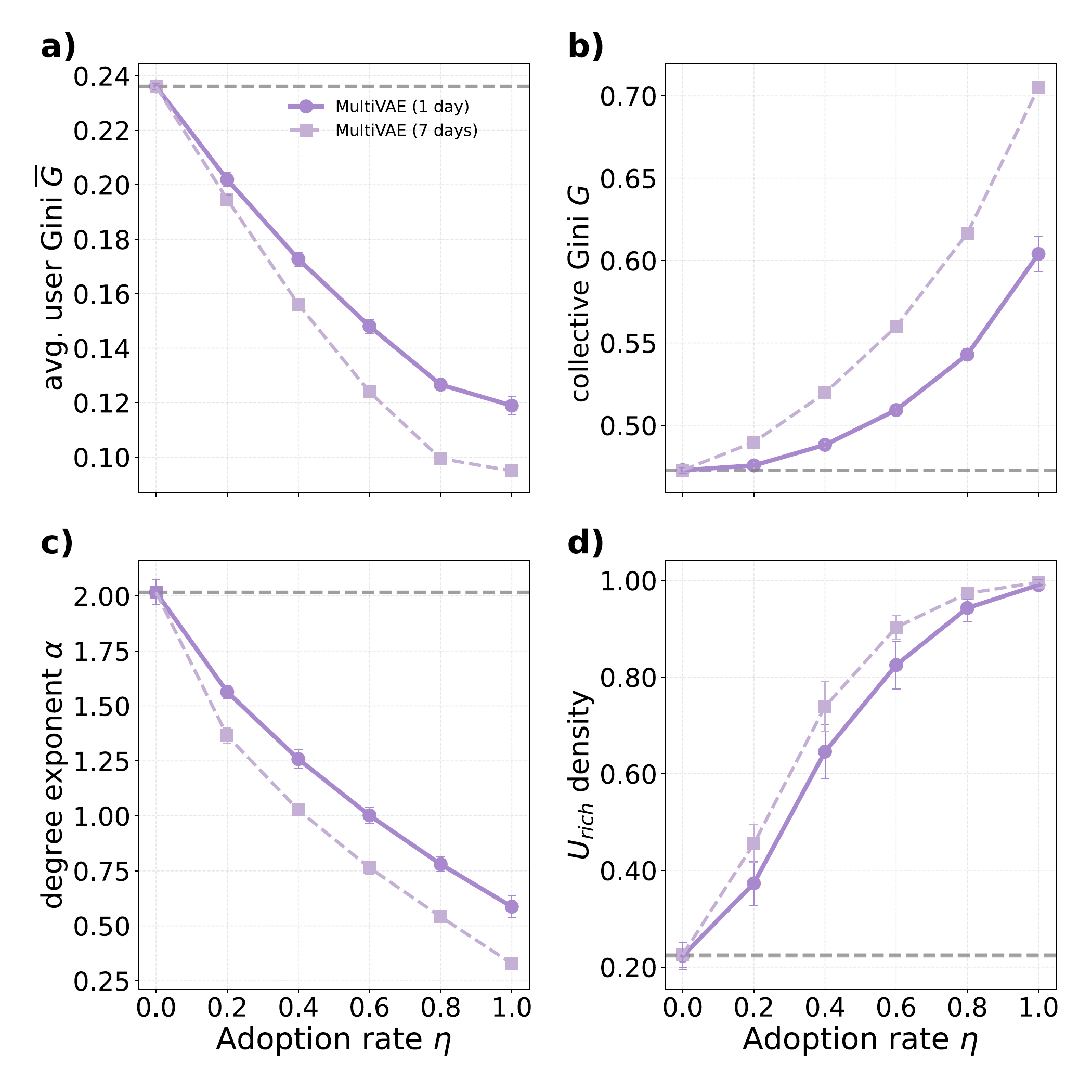}
    \caption{\textcolor{black}{Results obtained by comparing two different retraining frequencies: $\Delta = 1\; day$ and $\Delta = 7\; days$. No substantial differences are observed. The simulation uses MultiVAE as the recommender algorithm and the New York City dataset.}}
    \label{fig:shorter-retraining-freq}
\end{figure}

{\color{black}
\section{Impact of retraining frequency}
\label{appendix:generalization}
To assess the impact of the retraining frequency of recommender systems, we shorten the retraining interval of the simulation to $\Delta=1\; day$, i.e., we update the recommender system daily. 
As shown in Figure~\ref{fig:shorter-retraining-freq}, the results for $\Delta = 1$ day are consistent with those for $\Delta = 7$ days, although the effects are more pronounced in the latter case.
Our interpretation is that, because the simulation follows the real, longitudinal evolution of user–venue interactions, longer gaps between updates give the algorithm more time to amplify historical biases.}

\textcolor{black}{
We also replicate our experiments using a sliding-window training scheme instead of an incremental one.
In this setting, each new batch of simulated user–venue interactions replaces an equal number of the oldest interactions, keeping the training set size constant over time.
}

\begin{figure}
    \centering
    \includegraphics[width=0.75\linewidth]{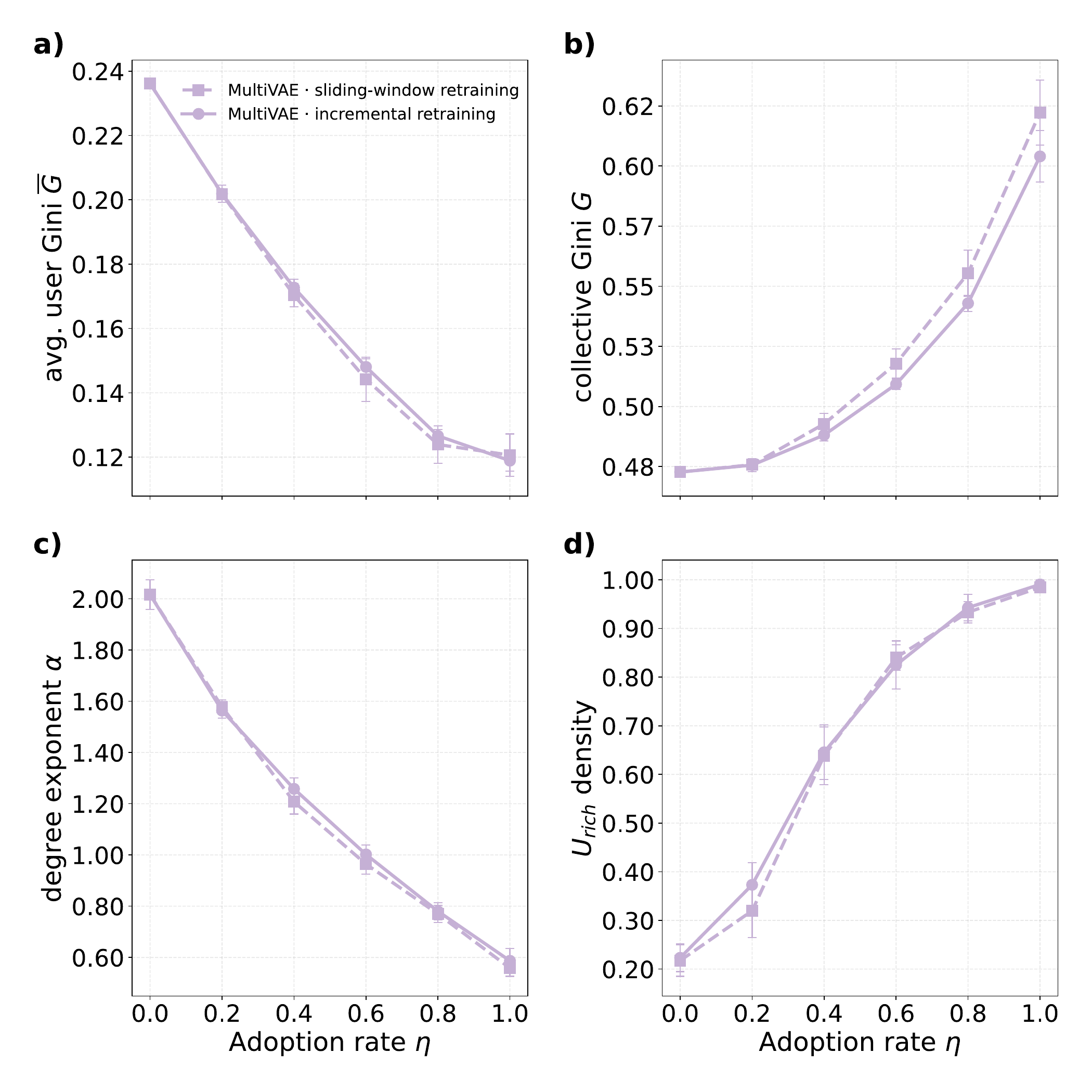}
    \caption{\textcolor{black}{Results obtained by comparing two training strategies: incremental training (used in the main text) and sliding-window training. No substantial differences are observed. The simulation uses MultiVAE as the recommender algorithm, the New York dataset, and a retraining interval of $\Delta = 1\; day$.}}
    \label{fig:sliding-window}
\end{figure}

\textcolor{black}{
As Figure \ref{fig:sliding-window} shows, the two schemas yield nearly identical outcomes: the curves mostly overlap and their error bars largely coincide, indicating no statistically significant differences. 
These findings suggest that our results are robust to the choice of training strategy, holding regardless of whether the recommender system is trained incrementally or using a sliding-window approach.
}

\section{Results for PGN}
\label{app:pgn}
\begin{figure}
    \centering
    \includegraphics[width=0.75\linewidth]{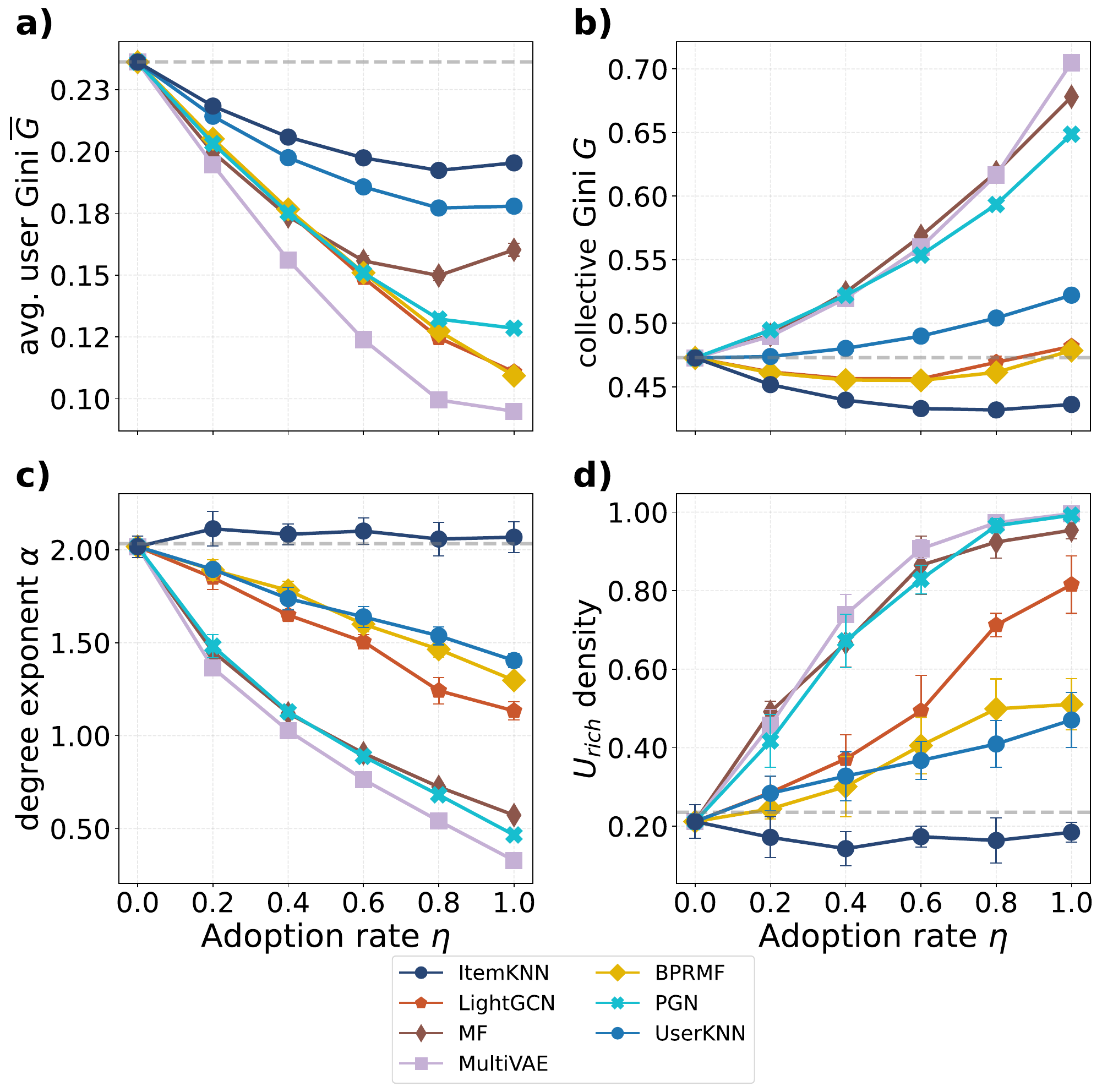}
    \caption{Effects of recommender system adoption $\eta$ considering also PGN.}
    \label{fig:app-with-pgn}
\end{figure}

Similarly to MF and MultiVAE, PGN leads to a notable reduction in the individual Gini coefficient, accompanied by a significant increase in the collective Gini coefficient (see Figure \ref{fig:app-with-pgn}).
We also observe a substantial decrease in the degree exponent $\alpha$ of the co-location network, suggesting a shift toward more homogenous connectivity, alongside the emergence of a denser rich-club structure.

\section{User similarity across adoption rates}
\label{appendix:similarity}

We study how user similarity evolves as a function of the adoption rate~$\eta$, offering further insight into the collective impact of different recommendation systems.
We define Jensen-Shannon similarity (JS similarity) as:
\[
\text{JS}(P, Q) = 1 - \mathrm{JS}(P \| Q)
\]
where $\mathrm{JS}(P \| Q)$ is the Jensen-Shannon divergence between two discrete probability distributions $P$ and $Q$. 
The JS divergence \citep{fuglede2004jensen} is defined as:
\[
\mathrm{JS}(P \| Q) = \frac{1}{2} \mathrm{KL}(P \| M) + \frac{1}{2} \mathrm{KL}(Q \| M)
\quad \text{where } M = \frac{1}{2}(P + Q)
\]
where $\mathrm{KL}$ denotes the Kullback–Leibler divergence \citep{kullback1997information, van2014renyi} defined as:
\[
\mathrm{KL}(P \| Q) = \sum_{x \in X} P(x) \log \left( \frac{P(x)}{Q(x)} \right)
\]

$\text{JS}(P, Q) \in [0, 1]$, it is symmetric and always well-defined.
The higher JS, the more similar the two distributions are. 

We select the top 50 most active users and the 50 most visited venues in New York City. For each user, we build a probability distribution over visits to these venues. We then compute the JS similarity between all user pairs at each value of the adoption rate~$\eta$.

\begin{figure}[h]
    \centering
    \begin{subfigure}[b]{0.48\textwidth}
        \includegraphics[width=\textwidth]{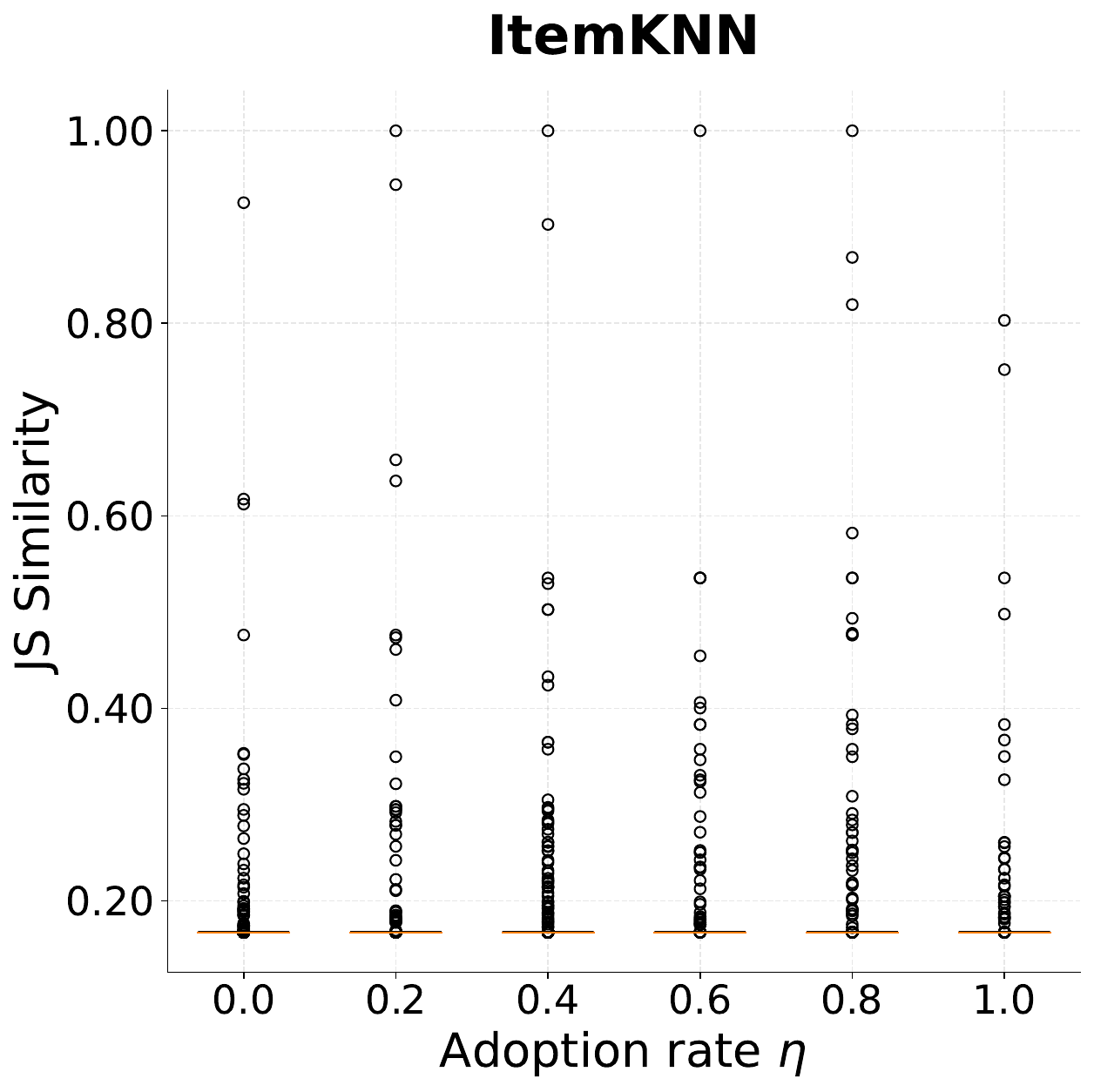}
        \caption{ItemKNN}
    \end{subfigure}
    \hfill
    \begin{subfigure}[b]{0.48\textwidth}
        \includegraphics[width=\textwidth]{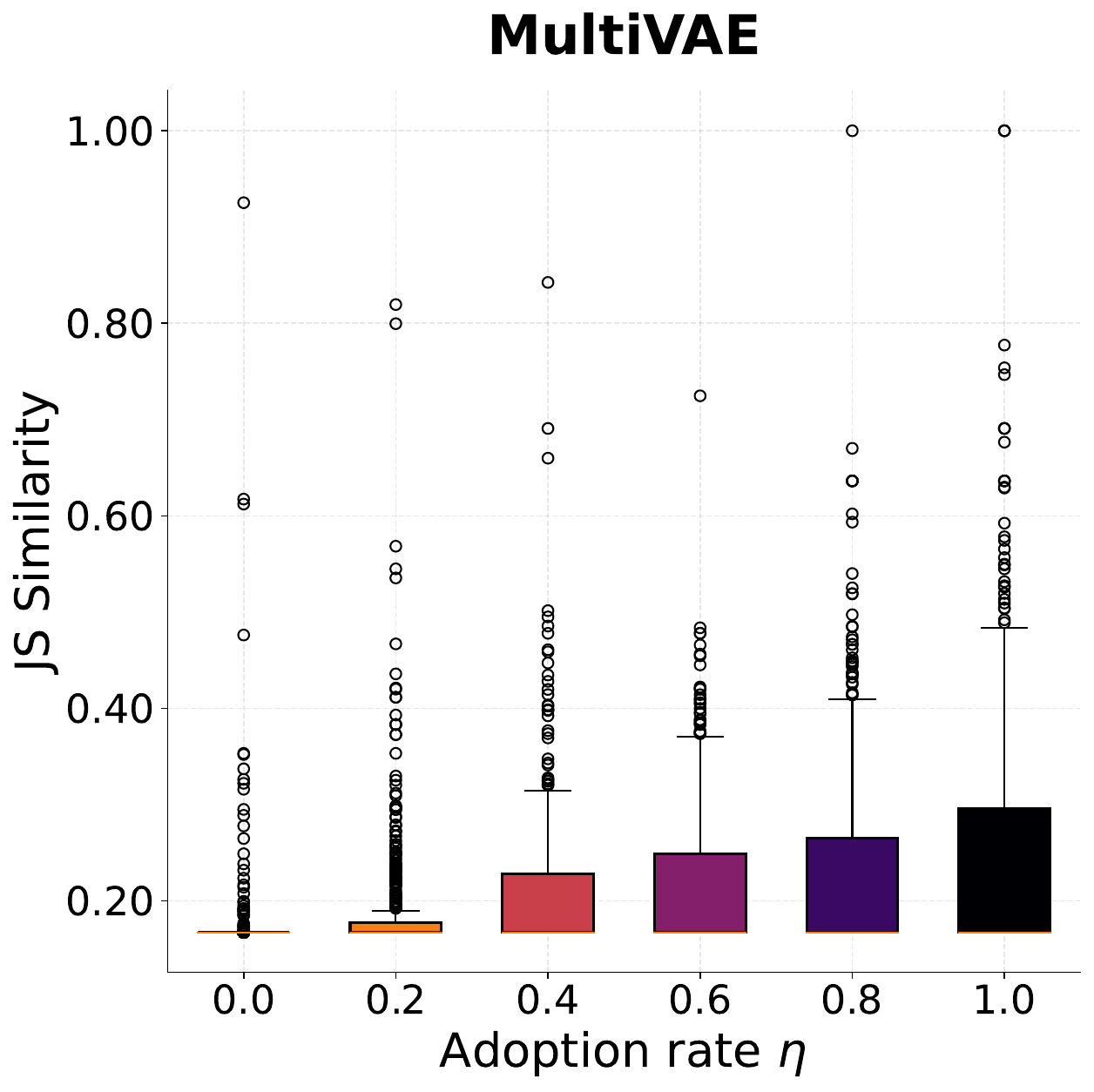}
        \caption{MultiVAE}
    \end{subfigure}
    \caption{Distribution of pairwise JS similarity among the top 50 users, computed over the 50 most visited venues, as a function of the adoption rate~$\eta$. Under ItemKNN, similarity remains relatively stable. In contrast, under MultiVAE, JS similarity increases with adoption, indicating that visitation patterns become more aligned as the recommender is more widely adopted.}
    \label{fig:similarity_boxplots}
\end{figure}

Figure~\ref{fig:similarity_boxplots} shows the Jensen-Shannon (JS) similarity as the adoption rate $\eta$ increases for ItemKNN and MultiVAE.
For MultiVAE, we observe a clear increase in JS similarity with rising $\eta$, indicating that higher reliance on the recommender system leads to more homogeneous visitation patterns.
In contrast, ItemKNN does not exhibit this trend.
These findings suggest that as users increasingly depend on MultiVAE for venue exploration, they become more similar to one another; they visit the same set of top venues.

